\documentclass[11pt]{article}
\usepackage[round]{natbib}

\usepackage{graphicx}
\usepackage{bm}
\usepackage{amssymb,amsmath}
\usepackage{amsthm}
\usepackage{comment}
\usepackage{dsfont}
\usepackage{fancyhdr}
\usepackage{algorithm, algpseudocode}
\usepackage{xr-hyper}
\usepackage[unicode,final]{hyperref}
\usepackage{mathtools}
\usepackage{multirow}
\usepackage{circledsteps}
\usepackage {cancel}
\usepackage{bibentry}
\usepackage[hang,small,bf]{caption}
\usepackage[subrefformat=parens]{subcaption}
\mathtoolsset{
  showonlyrefs,
  showmanualtags,
}

\newcommand{\argmin}{\mathop{\rm argmin}\limits}
\newcommand{\argmax}{\mathop{\rm argmax}\limits}

\newtheorem{assumption}{Assumption}
\newtheorem{proposition}{Proposition}
\newtheorem{corollary}{Corollary}

\allowdisplaybreaks  

\begin{document}

{\begin{flushleft}
{\LARGE Gradual Domain Adaptation via Normalizing Flows}
\end{flushleft}
\ \\
{\bf \large 
Shogo Sagawa$^{1}$ and
Hideitsu Hino$^{2,3}$ 
}

\begin{flushleft}
$^{1}$Department of Statistical Science,
School of Multidisciplinary Sciences, The Graduate University for Advanced Studies (SOKENDAI)\\
Shonan Village, Hayama, Kanagawa 240-0193, Japan\\
$^{2}$Department of Statistical Modeling, The Institute of Statistical Mathematics\\
10-3 Midori-cho, Tachikawa, Tokyo 190-8562, Japan\\
$^{3}$Center for Advanced Intelligence Project (AIP), RIKEN,\\
1-4-4 Nihonbashi, Chuo-ku, Tokyo 103-0027, Japan
\end{flushleft}}
{\bf Keywords: Gradual domain adaptation, Normalizing flow}
\thispagestyle{fancy}
\rhead{}
\lhead{}

\begin{center} {\bf Abstract} \end{center}
Standard domain adaptation methods do not work well when a large gap exists between the source and target domains. 
Gradual domain adaptation is one of the approaches used to address the problem. 
It involves leveraging the intermediate domain, which gradually shifts from the source domain to the target domain. 
In previous work, it is assumed that the number of intermediate domains is large and the distance between adjacent domains is small; hence, the gradual domain adaptation algorithm, involving self-training with unlabeled datasets, is applicable. 
In practice, however, gradual self-training will fail because the number of intermediate domains is limited and the distance between adjacent domains is large. 
We propose the use of normalizing flows to deal with this problem while maintaining the framework of unsupervised domain adaptation. 
The proposed method learns a transformation from the distribution of the target domain to the Gaussian mixture distribution via the source domain.
We evaluate our proposed method by experiments using real-world datasets and confirm that it mitigates the above-explained problem and improves the classification performance.

\section{Introduction}
In a standard problem of learning predictive models, it is assumed that the probability distributions of the test data and the training data are the same. 
The prediction performance generally deteriorates when this assumption does not hold. 
The simplest solution is to discard the training data and collect new samples from the distribution of test data.
However, this solution is inefficient and sometimes impossible, and there is a strong demand for utilizing valuable labeled data in the source domain.

Domain adaptation~\citep{ben2007analysis} is one of the transfer learning frameworks in which
the probability distributions of prediction target and training data are different. 
In domain adaptation, the source domain is a distribution with many labeled samples, and the target domain is a distribution with only a few or no labeled samples. 
The case with no labels from the target domain is called unsupervised domain adaptation and has been the subject of much research, including theoretical analysis and real-world application~\citep{ben2007analysis,cortes2010learning,mansour2009domain,redko2019advances,zhao2019learning}. 

In domain adaptation, the predictive performance on the target data deteriorates when the discrepancy between the source and target domains is large. 
\citet{kumar2020understanding} proposed gradual domain adaptation (GDA), in which it is assumed that the shift from the source domain to the target domain occurs gradually and that unlabeled datasets from intermediate domains are available. 
The key assumption of GDA~\citep{kumar2020understanding} is that there are many indexed intermediate domains.
The intermediate domains are arranged to connect the source domain to the target domain, and their order is known or {\it{index}} is given, starting from the one closest to the source domain to the one closer to the target domain. 
These intermediate domains connect the source and target domains densely so that gradual self-training is possible without the need for labeled data. 
In practice, however, the number of intermediate domains is limited. 
Therefore, the gaps between adjacent domains are large, and gradual self-training does not work well.  

\begin{figure}[!htbp]
\centering
  \includegraphics[clip, width=14cm]{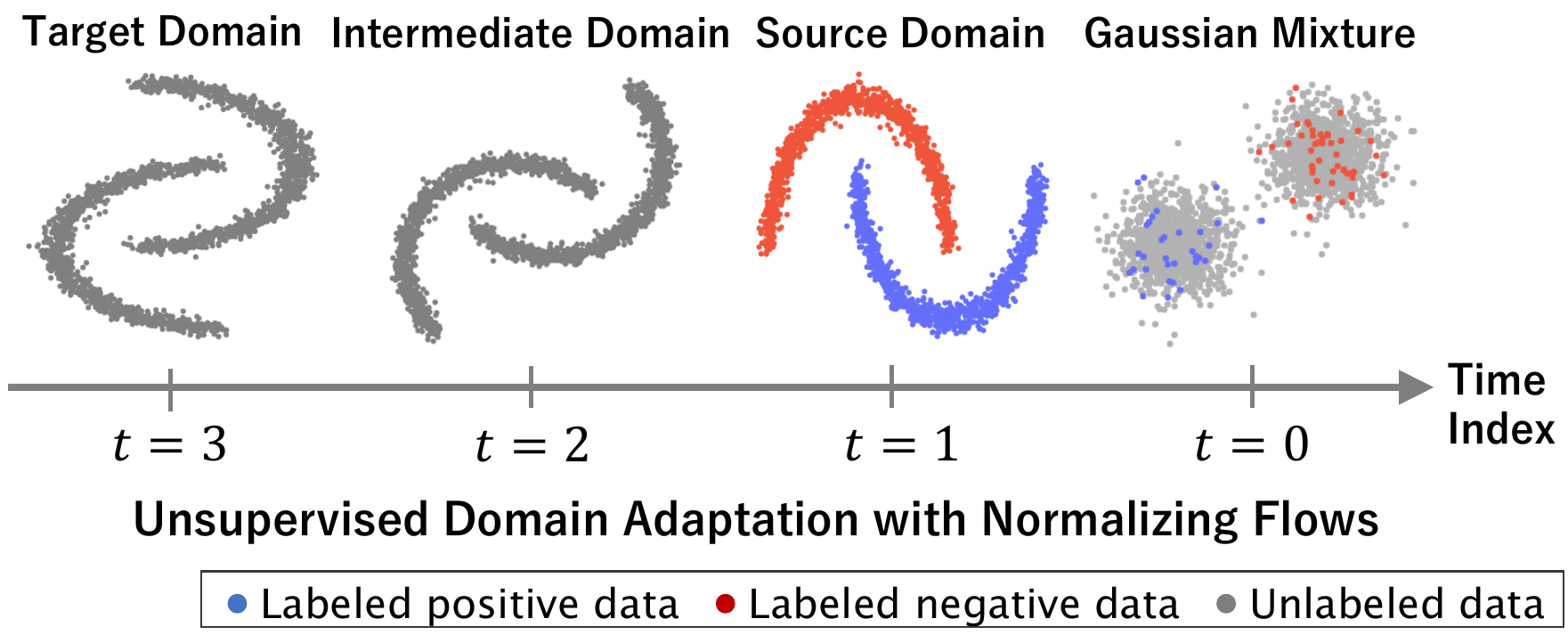}
  \caption{
  Overview of the proposed method. 
  Owing to the limited number of available intermediate domains, the applicability of gradual self-training is limited. 
  Gradual domain adaptation is possible without gradual self-training by utilizing continuous normalizing flow. 
  }
  \label{fig:overview}
\end{figure}

In this paper, we propose a method that mitigates the problem of large shifts between adjacent domains. 
Our key idea is to use generative models that learn continuous shifts between domains. 
Figure~\ref{fig:overview} shows a schematic of the proposed method. 
We focus on the normalizing flow (NF)~\citep{papamakarios2021normalizing} as a generative model to describe gradual shifts. 
We assume that the shifts in the transformation process correspond to gradual shifts between the source and target domains. 
Normalizing flows can achieve a more natural and direct transformation from the target domain to the source domain compared to other generative models, such as generative adversarial networks~\citep{pan2019recent}.
Inspired by the previous work~\citep{izmailov2020semi}, which utilizes an NF for semi-supervised learning, we propose a method to utilize an NF for gradual domain adaptation. 
Our trained NF predicts the class label of a sample from the target domain by transforming the sample to a sample from the Gaussian mixture distribution via the source domain. 
The transformation between the distribution of the source domain and the Gaussian mixture distribution is learned by leveraging labeled data from the source domain. 
Note that our proposed method does not use gradual self-training. 

The rest of the paper is organized as follows. 
We review related works in Section~\ref{sec:rw}. 
Then we explain in detail the gradual domain adaptation algorithm, an important previous work, in Section~\ref{sec:gda}. 
We introduce our proposed method in Section~\ref{sec:prop}. 
In Section~\ref{sec:exp}, we present experimental results. The last section is devoted to the conclusion of our study.

\section{Related works}\label{sec:rw}
We tackle the gradual domain adaptation problem by using the normalizing flow. 
These topics have been actively researched in recent years, making it challenging to provide a comprehensive review. 
Here, we will introduce a few closely related studies.

\subsection{Gradual domain adaptation}
In conventional domain adaptation, a model learns the direct transformation between (samples from) the source and target domains. 
Several methods have been proposed to transfer the source domain to the target domain sequentially~\citep{gadermayr2018gradual, gong2019dlow, hsu2020progressive, choi2020visual, cui2020gradually, dai2021idm}. 
A sequential domain adaptation is realized by using data generated by mixing the data from the source and target domains.

\citet{kumar2020understanding} proposed gradual domain adaptation (GDA), and they showed that it is possible to adapt the method to a large domain gap by self-training with unlabeled datasets. 
It is assumed that the intermediate domains gradually shift from the source domain to the target domain, and the sequence of intermediate domains is given. 
\citet{chen2021gradual} developed the method in which the intermediate domains are available whereas their indices are unknown. 
\citet{zhang2021gradual} and \citet{abnar2021gradual} proposed to apply the idea of GDA to conventional domain adaptation.
Since in these methods, they assumed that intermediate domains are unavailable, they use pseudo-intermediate domains. 
\citet{zhou2022active} proposed a gradual semi-supervised domain adaptation method that utilizes self-training and requests of queries. 
They also provided a new dataset suitable for GDA. 
\citet{kumar2020understanding} conducted a theoretical analysis and provided a generalization error bound for gradual self-training. 
\citet{pmlr-v162-wang22n} conducted a theoretical analysis under more general assumptions and derived an improved generalization error bound. 
\citet{dong2022algorithms} also conducted a theoretical analysis under the condition that all the labels of the intermediate domains are given. 
\citet{he2023gradual} assumed a scenario where, similar to ours, the available intermediate domains are limited. 
They propose generating a pseudo-intermediate domain using optimal transport and use self-training to propagate the label information of the source domain.

There are several problem settings similar to those in GDA. 
\citet{NEURIPS2020_fd69dbe2} proposed evolving domain adaptation, demonstrating the feasibility of adapting a target domain that evolves over time through meta-learning. 
In contrast to GDA, which has only one target domain, the evolving domain adaptation assumes that the sequence of the target domains is given and aims at achieving accurate prediction over all the target domains.
\citet{pmlr-v119-wang20h} assumed the problem where there are multiple source domains with indices, which corresponds to the GDA problem where all labels of the intermediate domains can be accessed. 
\citet{huang2022curriculum} proposed the application of the idea of GDA to reinforcement learning. 
They propose a method, following a similar concept to curriculum learning~\citep{10.1145/1553374.1553380}, that starts with simple tasks and gradually introducing more challenging problems for learning. 

Multi-source domain adaptation~\citep{zhao2020multi} corresponds to GDA under the condition that all labels of the intermediate domains are given while the indices of the intermediate domains are not given. 
\citet{pmlr-v180-ye22b} proposed a method for temporal domain generalization in online recommendation models. 
\citet{zhou2022online} proposed an online learning method that utilizes self-training and requests of queries. 
\citet{pmlr-v180-ye22b} and \citet{zhou2022online} assumed that the labels of the intermediate domains are given, while \citet{SAGAWA2023731} applied the multifidelity active learning assuming access to the labels of the intermediate and target domains at certain costs.

\subsection{Normalizing flows}
Normalizing flows (NFs) are reversible generative models that use invertible neural networks to transform samples from a known distribution, such as the Gaussian distribution. 
NFs are trained by the maximum likelihood estimation, where the probability density of the transformed random variable is subject to the change of variable formula. 
The architecture of the invertible neural networks is constrained (e.g., coupling-based architecture) so that its Jacobian matrix is efficiently computed.
NFs with constrained architectures are called discrete NFs (DNFs), and examples include RealNVP~\citep{DBLP:conf/iclr/DinhSB17}, Glow~\citep{kingma2018glow}, and Flow++~\citep{ho2019flow++}. 
Several theoretical analyses of the expressive power of DNFs have also been reported. 
\citet{kong2020expressive} studied basic flow models such as planar flows~\citep{rezende2015variational} and proved the bounds of the expressive power of basic flow models. 
\citet{teshima2020coupling} conducted a more generalized theoretical analysis of coupling-based flow models. 
\citet{chen2018neural} proposed continuous normalizing flow, mitigating the constraints on the architecture of the invertible neural networks.
CNFs describe the transformation between samples from the Gaussian to the observed samples from a complicated distribution using ordinary differential equations. 
\citet{grathwohl2018ffjord} proposed a variant of CNF called FFJORD, which exhibited improved performance over DNF. 
FFJORD was followed by studies to improve the computational efficiency~\citep{huang2021accelerating,onken2021ot} and on the representation on manifolds~\citep{mathieu2020riemannian,rozen2021moser,ben2022matching}.

\citet{brehmer2020flows} suggested that NFs are unsuitable for data that do not populate the entire ambient space. 
Several normalizing flow models aiming to learn a low-dimensional manifold on which data are distributed and to estimate the density on that manifold have been proposed~\citep{brehmer2020flows,caterini2021rectangular,horvat2021denoising,kalatzis2021density,ross2021tractable}.

Normalizing flows have been applied to several specific tasks, for example, data generation  \lbrack images~\citep{lu2020structured}, 3D point clouds~\citep{pumarola2020c}, chemical graphs~\citep{kuznetsov2021molgrow}\rbrack, anomaly detection~\citep{kirichenko2020normalizing}, and semi-supervised learning~\citep{izmailov2020semi}. 
NFs are also used to compensate for other generative models~\citep{mahajan2019latent, yang2019pointflow,abdal2021styleflow, ijcai2021-103} such as generative adversarial networks and variational autoencoders (VAEs)~\citep{8616075}. 

As an approach of domain adaptation, it is natural to learn domain-invariant representations between the source and target domains, and several methods that utilize NFs have been developed for that purpose. 
\citet{grover2020alignflow} and \citet{das2021cdcgen} proposed a domain adaptation method that combines adversarial training and NFs. 
These methods separately train two NFs for the source and target domains and execute domain alignment in a common latent space using adversarial discriminators. 
\citet{askari2023mapflow} proposed a domain adaptation method using NFs and VAEs. 
The encoder converts samples from the source and target domains into latent variables. 
The latent space of the source domain is forced to be Gaussian. 
NFs transform latent variables of the target domain into latent variables of the source domain and predict the class label of the target data.
The above-explained methods do not assume a situation where there is a large discrepancy between the source and target domains, as assumed in GDA.

\section{Formulation of gradual domain adaptation}\label{sec:gda}
In this section, we introduce the concept and formulation of the gradual domain adaptation proposed by~\citet{kumar2020understanding}, which utilizes gradual self-training, and we confirm that a gradual self-training-based method is unsuitable when the discrepancies between adjacent domains are large. 

Consider a multiclass classification problem. 
Let $\mathcal{X} = \mathbb{R}^{d}$ and $\mathcal{Y} = \{1, 2, \ldots, C\}$ be the input and label spaces, respectively. 
The source dataset has labels $S=\{(\bm{x}_i^{(1)}, y_i^{(1)})\}_{i=1}^{n_{1}}$, whereas the intermediate datasets and the target dataset do not have labels $U^{(j)}=\{\bm{x}_i^{(j)}\}_{i=1}^{n_{j}}$. 
The subscript $i, \; 1 \leq i \leq n_j$ indicates the $i$-th observed datum, and the superscript $(j), \; 1 \leq j \leq K$ indicates the $j$-th domain. 
We note that $\{U^{(j)}\}_{j=2}^{K-1}$ are the intermediate datasets and $U^{(K)}$ is the target dataset. 
The source domain corresponds to $j=1$, and the target domain corresponds to $j=K$. 
When $j$ is small, the domain is considered to be similar to the source domain. 
In contrast, when $j$ is large, the domain is considered to be similar to the target domain. 
Let $p_j$ be the probability density function of the $j$-th domain.
The Wasserstein metrics~\citep{villani2009optimal} are used to measure the distance between domains. 
\citet{kumar2020understanding} defined the distance between adjacent domains as the per-class $\infty$-Wasserstein distance. 
\citet{pmlr-v162-wang22n} defined the distance between adjacent domains as the $p$-Wasserstein distance as a more general metric. 
Following~\citet{pmlr-v162-wang22n}, we define the average $p$-Wasserstein distance between consecutive domains as $\rho = \frac{1}{K-1} \sum_{j=2}^{K} W_p(p_{j-1}(\bm{x},y), p_{j}(\bm{x},y))$, where $W_p(\cdot,\cdot)$ denotes the $p$-Wasserstein distance. 

\citet{kumar2020understanding} proposed a GDA algorithm that consists of two steps. 
In the first step, the predictive model for the source domain is trained with the source dataset. 
Then, by sequential application of the self-training, labels of the adjacent domains are predicted. 
Let $\mathcal{H} = \{h \mid h: \mathcal{X} \to \mathcal{Y}\}$ and $\ell: \mathcal{Y} \times \mathcal{Y} \to \mathbb{R}_{\geq 0}$ be a hypothesis space and a loss function, respectively. 
We consider training the model $h^{(1)}$ by minimizing the loss on the source dataset 
\begin{equation}
    h^{(1)} = \argmin_{h \in \mathcal{H}} \frac{1}{n_{1}} \sum_{i=1}^{n_{1}} \ell(h(\bm{x}_i^{(1)}), y_i^{(1)}).
    \label{eq:base-train}
\end{equation}
For the joint distribution $p_{j}(\bm{x},y)$, the expected loss with the classifier $h^{(j)}$ is defined as $\epsilon^{(j)}(h^{(j)}) = \mathbb{E}_{\bm{x},y \sim p_{j}(\bm{x},y)}[\ell(h^{(j)}(\bm{x}),y)]$. 

The classifier of the current domain $h^{(j)}$ is used to make predictions on the unlabeled dataset $U^{(j+1)}=\{\bm{x}_i^{(j+1)}\}_{i=1}^{n_{j+1}}$ in the next domain. 
Let $\mathrm{ST}(h^{(j)}, U^{(j+1)})$ be a function that returns the self-trained model for $\bm{x}^{(j+1)} \in U^{(j+1)}$ by inputting the current model $h^{(j)}$ and an unlabeled dataset $U^{(j+1)}$: 
\begin{align}
    \mathrm{ST}(h^{(j)}, U^{(j+1)})
    = \argmin_{h \in \mathcal{H}} \frac{1}{n_{j+1}} \sum_{i=1}^{n_{j+1}} \ell(h(\bm{x}_{i}^{(j+1)}),h^{(j)}(\bm{x}_{i}^{(j+1)})).
    \label{eq:self-train}
\end{align}
Self-training is only applied between adjacent domains. 
The output of GDA is the classifier for the target domain $h^{(K)}$. 
The classifier $h^{(K)}$ is obtained by applying sequential self-training to the model of the source domain $h^{(1)}$ along the sequence of unlabeled datasets $U^{(2)}, \ldots, U^{(K)}$ denoted as $
h^{(2)} \!=\! \mathrm{ST}(h^{(1)}, U^{(2)}),
\cdots, 
h^{(K)} \!=\! \mathrm{ST}(h^{(K-1)}, U^{(K)}).
$

\citet{kumar2020understanding} provided the first generalization error bound for the gradual self-training of $e^{\mathcal{O}(K-1)}(\epsilon^{(1)}(h^{(1)}) \!+\! \mathcal{O}(\sqrt{\frac{\log K-1}{n}}))$, where the sample size in each domain is assumed to be the same: $n_2\!=\! \cdots \!=\!n_K\!=\!n$ without loss of generality. 
\citet{pmlr-v162-wang22n} conducted a theoretical analysis under more general assumptions and derived an improved generalization bound: 
\begin{equation}
    \epsilon^{(1)}(h^{(1)})\!+\!\tilde{\mathcal{O}}\left(\rho\cdot (K\!-\!1)\!+\!\frac{K\!-\!1}{\sqrt{n}}\!+\!\frac{1}{\sqrt{n(K\!-\!1)}}\right). 
    \label{eq:bounds}
\end{equation}
When we consider the problem where the number of accessible intermediate domains is limited, it is natural that the distance between adjacent domains $\rho$ becomes large. 
When this occurs, Eq.~\eqref{eq:bounds} suggests that the bound for the expected loss of the target classifier will become loose. 
To tackle the problem of a large distance between adjacent domains, we propose a method utilizing generative models.

\section{Proposed method}\label{sec:prop}
We consider a GDA problem with large discrepancies between adjacent domains. 
As discussed in Section~\ref{sec:gda}, the applicability of a gradual self-training-based methods~\citep{kumar2020understanding,pmlr-v162-wang22n,he2023gradual} is limited in this situation. 
Our proposed GDA method mitigates the problem without gradual self-training. 
Our key idea is to utilize NFs to learn gradual shifts between domains as detailed in Section~\ref{sec:prop-cnf}. 
Conventional NFs only learn the transformation between an unknown distribution and the standard Gaussian distribution, whereas GDA requires transformation between unknown distributions. 
Section~\ref{sec:prop-knn} introduces a nonparametric likelihood estimator and shows how the likelihood of transformation between samples from adjacent domains is evaluated. 
We consider a multiclass classification problem; hence, our flow-based model learns the transformation between the distribution of the source domain and a Gaussian mixture distribution as detailed in Section~\ref{sec:prop-gm}. 
We discuss the theoretical aspects and the scalability of the proposed method in Sections~\ref{sec:prop-theory} and~\ref{sec:prop-limit}, respectively.

\subsection{Learning gradual shifts with normalizing flows}\label{sec:prop-cnf}
An NF uses an invertible function $f:\mathbb{R}^{d}\to\mathbb{R}^{d}$ to transform a sample $\bm{x} \in \mathbb{R}^{d}$ from the complicated distribution $p(\bm{x})$ to a sample $\bm{z} \in \mathbb{R}^{d}$ from the standard Gaussian $p_0(\bm{z})$.
The log density of $\bm{x} = f(\bm{z})$ satisfies
$
\log p(\bm{x}) 
= \log p_0(f^{-1}(\bm{x}))
+ \log \left|\mathrm{det} \nabla f^{-1}(\bm{x})\right|,
$
where $\nabla f^{-1}(\bm{x})$ is the Jacobian of $f^{-1}$. 

Our aim is to learn the continuous change $\bm{x}^{(K)} \!\mapsto\!\cdots\!\mapsto\!\bm{x}^{(1)}\!\mapsto\!\bm{z}$ by using NFs. 
We consider a continuous transformation, 
$
\bm{x}^{(K)}\!=\!f^{(K)}(\bm{x}^{(K-1)}),
\ldots,
\bm{x}^{(1)}\!=\!f^{(1)}(\bm{z})
$, by using multiple NFs $f^{(K)}, \ldots, f^{(1)}$.
Our preliminary experiments indicate that continuous normalizing flows (CNFs) are better suited for capturing continuous transitions between domains than discrete normalizing flows. 
Further details of these experiments are elaborated upon in Section~\ref{sec:sub-exp-dnf}. 

To learn gradual shifts between domains with CNFs, we regard the index of each domain $j$ as a continuous variable. 
We introduce a time index $t \in \mathbb{R}_{+}$ to represent continuous changes of domains, and the index of each domain $j$ is considered as a particular time point. 
The probability density function $p_j$ is a special case of $p_t$ when $t\!=\!j$.
The CNF $g: \mathbb{R}^{d} \times \mathbb{R}_{+} \to \mathbb{R}^{d}$ outputs a transformed variable dependent on time $t$, and we consider $f^{(t)}(\cdot)$ as $g(\cdot, t)$ following the standard notation of CNF. 
We set $t\!=\!0$ and $t\!=\!j$ for $\bm{z}$ and $\bm{x}^{(j)}$, respectively. 
Note that $\bm{z} = g(\bm{z}, 0)$ and $\bm{x}^{(j)} = g(\bm{x}^{(j)}, j)$. 
Let $v$ be a neural network parameterized by $\omega$ that represents the change in $g$ along $t$. 
Following~\citet{chen2018neural} and \citet{grathwohl2018ffjord}, we express the ordinary differential equation (ODE) with respect to $g$ using the neural network $v$ as $\partial g/ \partial t = v(g(\cdot, t), t; \omega)$. 
The parameter $\omega$ of the neural network $v$ implicitly defines the CNF $g$. 
When we specifically refer to the parameter of $g$, we denote it as $g_\omega$. 

An NF requires an explicit computation of the Jacobian, and in a CNF, it is calculated by integrating the time derivative of the log-likelihood. 
The time derivative of the log-likelihood is expressed as $\partial \log p(g) / \partial t = -\mathrm{Tr}(\partial v / \partial g)$~\citep[Theorem 1]{chen2018neural}.
The outputs of the CNF are acquired by solving an initial value problem. 
Since our goal is to learn the gradual shifts between domains using CNFs, we consider solving the initial value problem sequentially. 
To formulate the problem, we introduce the function $\tau$ that relates the time index $t$ to the input variable as follows: 
\begin{align}
    \tau(t) = 
    \begin{dcases*}
        \bm{x}^{(t)}, & if $t \geq 1$,\\
        \bm{z}, & otherwise. 
    \end{dcases*}
\end{align}
We assign $t_0\!=\!j-1$ and $t_1\!=\!j$, and the output of the CNF for the input $\bm{x}^{(j)}$ is obtained by solving the following initial value problem: 
\begin{align}
\begin{split}
    \begin{bmatrix}
       \tau(t_0) \\
       \Delta_1
    \end{bmatrix}
    &\!=\! \int_{t_1}^{t_0}
    \begin{bmatrix}
       v(g(\bm{x}^{(j)}, t),t;\omega) \\
       -\mathrm{Tr} \left( \frac{\partial v}{\partial g} \right)
    \end{bmatrix} dt
    ,\\
    \begin{bmatrix}
       g(\bm{x}^{(j)}, t_1) \\
       \Delta_0
    \end{bmatrix}
    &\!=\!
    \begin{bmatrix}
       \bm{x}^{(j)} \\
       0
    \end{bmatrix}
    \label{eq:solveODE}
\end{split}
\end{align}
where $\Delta_1\!=\!\log p_{t_1}(\tau(t_1)) - \log p_{t_0}(\tau(t_0))$ and $\Delta_0\!=\!\log p_{t_1}(\tau(t_1)) - \log p_{t_1}(g(\bm{x}^{(j)},t_1))$.
To generate the shift between multiple domains, when $j>1$, the initial value problem is solved sequentially with decreasing values of $t_1$ and $t_0$ until $t_1=1$ and $t_0=0$. 
For instance, when $j\!=\!2$, we solve the initial value problem given by Eq.~\eqref{eq:solveODE} and retain the solutions. 
In the next iteration, we decrease the values of $t_1$ and $t_0$ and utilize the retained values as initial values. 
A CNF is formulated as a problem of maximizing the following log-likelihood with respect to the parameter $\omega$:
\begin{align}
    \log p_{j}(g_{\omega}(\bm{x}^{(j)},j)) 
    = \sum_{t=1}^{j} \log p_{t-1}(g_{\omega}(\bm{x}^{(j)}, t\!-\!1))
    - \int_{0}^{j} \mathrm{Tr} \left(\frac{\partial v}{\partial g_{\omega}}\right) dt.
    \label{eq:cnf}
\end{align}

\subsection{Non-parametric estimation of log-likelihood}\label{sec:prop-knn}
The transformation of a sample from the $j$-th domain to a sample from the adjacent domain using a CNF requires the computation of the log-likelihood $\log p_{t-1}(g_{\omega}(\bm{x}^{(t)}, t\!-\!1))$, where $t\!=\!j$. 
We use a $k$ nearest neighbor ($k$NN) estimators for the log-likelihood~\citep{Kozachenko198795,doi:10.1080/104852504200026815}. 
We compute the Euclidean distance between all samples in $\{\bm{x}_i^{(t-1)}\}_{i=1}^{n_{t-1}}$ and $g_{\omega}(\bm{x}^{(t)}, t\!-\!1)$, with $g_{\omega}(\bm{x}^{(t)}, t\!-\!1)$ kept fixed. 
Let $\delta_{t-1}^{k}(g_{\omega}(\bm{x}^{(t)}, t\!-\!1))$ be the Euclidean distance between $g_{\omega}(\bm{x}^{(t)}, t\!-\!1)$ and its $k$-th nearest neighbor in $\{\bm{x}_i^{(t-1)}\}_{i=1}^{n_{t-1}}$. 
The log-likelihood of the sample $g_{\omega}(\bm{x}^{(t)}, t\!-\!1)$ is estimated as
\begin{align}
    \log p_{t-\!1}(g_{\omega}(\bm{x}^{(t)}, t\!-\!1)) 
    \propto
    \!- d \log \delta_{t-\!1}^{k}(g_{\omega}(\bm{x}^{(t)}, t\!-\!1)).  
    \label{eq:knn}
\end{align}
When training our flow-based model, we estimate the log-likelihood using Eq.~\eqref{eq:knn} for all samples in the $t$-th domain $\{\bm{x}_i^{(t)}\}_{i=1}^{n_{t}}$ and minimize its sample average $-\frac{d}{n_{t}} \sum_{i=1}^{n_{t}} \log \delta_{t-1}^{k}(g_{\omega}(\bm{x}_{i}^{(t)}, t\!-\!1)))$ with respect to $\omega$.

For simplicity, we consider the case $n_{t}\!=\!n_{t-1}\!=\!n$. 
The cost of computing the log-likelihood by the $k$NN estimators is $\mathcal{O}(n^2)$. 
During the training of CNF $g_{\omega}$, the computation of log-likelihood by the $k$NN estimators is required each time CNF $g_{\omega}$ is updated. 
We use the nearest neighbor descent algorithm~\citep{dong2011efficient} to reduce the computational cost of the $k$NN estimators. 
The algorithm can be used to construct $k$NN graphs efficiently, and the computational cost is empirically evaluated to be $\mathcal{O}(n^{1.14})$. 

Another way to estimate the log-likelihood $\log p_{t-1}(g(\bm{x}^{(t)}, t\!-\!1))$ is to approximate $p_{t-1}$ using a surrogate function. 
Our preliminary experiments show that the $k$NN estimators are suitable for learning continuous changes between domains, and the details of the experiments are described in Section~\ref{sec:sub-exp-fitted_gmm}.

\subsection{Gaussian mixture model}\label{sec:prop-gm}
An NF transforms observed samples into samples from a known probability distribution. 
\citet{izmailov2020semi} proposed a semi-supervised learning method with DNFs using a Gaussian mixture distribution as the known probability distribution. 
In this subsection, we explain the Gaussian mixture model (GMM) suitable for our proposed method and the log-likelihood of the flow-based model with respect to the GMM.

The distribution $p_0$, conditioned on the label $s$, is modeled by a Gaussian with the mean $\bm{\mu}_s$ and the covariance matrix $\Sigma_s$, $p_0(\bm{z}|y\!=\!s) \!=\! \mathcal{N}(\bm{z}|\bm{\mu}_s, \Sigma_s)$. 
Following~\citet{izmailov2020semi}, we assume that the classes $\{1, 2, \ldots, C\}$ are balanced, i.e., $\forall s \!\in\! \{1,2,\dots,C\},\;p(y\!=\!s)\!=\!1/C$, and the Gaussian mixture distribution is $p_0(\bm{z}) \!=\! \frac{1}{C} \sum_{s=1}^{C} \mathcal{N}(\bm{z}|\bm{\mu}_s, \Sigma_s)$. 
The Gaussian for different labels should be distinguishable from each other, and it is desirable that the appropriate mean $\bm{\mu}_s$ and the covariance matrix $\Sigma_s$ are assigned to each Gaussian distribution. 
We set an identity matrix as the covariance matrix for all classes, $\Sigma_s\!=\!I$. 
We propose to assign the mean vector $\bm{\mu}_s = [\mu_{s_1}, \ldots, \mu_{s_d}]^\top$ using the polar coordinates system. 
Each component of the mean vector is given by
\begin{equation}
    \mu_{s_i} =
    \begin{cases}
        \displaystyle
        r\cos\theta_s (\sin\theta_s)^{i-1}, &(i=1,\ldots,d-1), \\
        r(\sin\theta_s)^{d-1}, &(i=d),
    \end{cases} 
    \label{eq:mean_r} \noeqref{eq:mean_r}
\end{equation}
where $r$ is the distance from the origin in the polar coordinate system and the angle $\theta_s\!=\!2 \pi (s\!-\!1) / C, \forall s \in \{1,2,\dots,C\}$. Note that $r$ is a hyperparameter. 

Since the source domain has labeled data, by using Eq.~\eqref{eq:cnf}, we can obtain the class conditional log-likelihood of a labeled sample as
\begin{align}
    \log p_{1}(g_{\omega}(\bm{x}^{(1)},1)|y=s) 
    = \log\mathcal{N}(g_{\omega}(\bm{x}^{(1)},0)|\bm{\mu}_s, \Sigma_s)
    -\int_{0}^{1}\mathrm{Tr} \left(\frac{\partial v}{\partial g_{\omega}}\right)dt.\label{eq:ll-label}
\end{align}
The intermediate domains and the target domain have no labeled data. The log-likelihood of an unlabeled sample is given by
\begin{align}
    \log p_{j}(g_{\omega}(\bm{x}^{(j)},j)) 
    &=\log\!\left\{\frac{1}{C}\sum_{s=1}^{C}\mathcal{N}(g_{\omega}(\bm{x}^{(j)},0)|\bm{\mu}_s,\Sigma_s)\!\right\} \nonumber \\
    &- \sum_{t=2}^{j} d\log \delta_{t-1}^{k}(g_{\omega}(\bm{x}^{(j)},t\!-\!1))
    - \int_{0}^{j} \mathrm{Tr}\left(\frac{\partial v}{\partial g_{\omega}}\right) dt.
    \label{eq:ll-unlabel} \noeqref{eq:ll-unlabel}
\end{align}
Namely, to learn the gradual shifts between domains, we maximize the log-likelihood of our flow-based model $g_{\omega}$ on all the data from the initially given domains. The algorithm minimizes the following objective function with respect to the flow-based model $g_{\omega}$:
\begin{align*}
    \mathrm{L}(\omega;S, \{U^{(j)}\}_{j=2}^{K}) 
    = - \frac{1}{n_1} \sum_{i=1}^{n_1}\log p_{1}(g_{\omega}(\bm{x}_{i}^{(1)},1)|y_i)
    - \sum_{j=2}^{K} \frac{1}{n_j} \sum_{i=1}^{n_j}\log p_{j}(g_{\omega}(\bm{x}_{i}^{(j)}, j)).
\end{align*}
We show a pseudocode for our proposed method in Algorithm~\ref{fullalgo}. 

Our method has two hyperparameters, $k$ and $r$: $k$ affects the computation of log-likelihood by $k$NN estimators, whereas $r$ controls the distance between the Gaussian corresponding to each class. 
We discuss how to tune these hyperparameters in Section~\ref{sec:sub-exp-param}. 

We consider making predictions for a new sample by using our flow-based model. 
The predictive probability that the class of the given test sample $\bm{x}$ being $s$ is
\begin{align}
    p(y\!=\!s|\bm{x}) 
    \!=\! \frac{p(\bm{x}|y\!=\!s)p(y\!=\!s)}{\sum_{s'=1}^{C} p(\bm{x}|y\!=\!s')p(y\!=\!s')}
    \!=\! \frac{\mathcal{N}(g_{\omega}(\bm{x},0)|\mu_s,\Sigma_s)}{\sum_{s'=1}^{C} \mathcal{N}(g_{\omega}(\bm{x},0)|\mu_{s'},\Sigma_{s'})}.
    \label{eq:predict}
\end{align}
Therefore, the class label of a new sample $\bm{x}$ is predicted by
\begin{equation}
    \hat{y} = \argmax_{s \in \{1, \ldots, C\}} p(y=s|\bm{x}).
\end{equation}

\begin{algorithm}[!htbp]
\caption{Gradual Domain Adaptation with CNF}
\label{fullalgo}
\begin{algorithmic}[1]
\renewcommand{\algorithmicrequire}{\textbf{Input:}}
\renewcommand{\algorithmicensure}{\textbf{Output:}}
\algnewcommand\algorithmicforeach{\textbf{for each}}
\algdef{S}[FOR]{ForEach}[1]{\algorithmicforeach\ #1\ \algorithmicdo}
{\scriptsize
\Require labeled dataset $S$ and unlabeled datasets $U^{(2)},\ldots,U^{(K)}$
\Ensure trained CNF $g_{\omega}$
\State $j \leftarrow K$
\Comment{start training from the target domain}
\While{$j > 0$}
    \State $t_0 \leftarrow j-1$, \hspace{1mm} $t_1 \leftarrow j$.
    \State initial values are set to $\bm{x}^{(j)}$ and $0$.
    \While{$t_0 > 0$}
    \Comment{the initial value problem is solved sequentially}
        \State solve the initial value problem Eq.~\eqref{eq:solveODE} to obtain $g_{\omega}(\bm{x}^{(j)}, t_0)$ and $-\mathrm{Tr} \left( \frac{\partial v}{\partial g} \right)$.
        \State retain the solutions.
        \Comment{these values will be used as initial values in the next iteration}
        \State $t_0 \leftarrow t_0-1$, \hspace{1mm} $t_1 \leftarrow t_1-1$.
    \EndWhile
    \If{$j = 1$}
    \Comment{update CNF $g_{\omega}$}
        \State maximize the log-likelihood of labeled data as Eq.~\eqref{eq:ll-label} with respect to $\omega$.
    \Else
        \State maximize the log-likelihood of unlabeled data as Eq.~\eqref{eq:ll-unlabel} with respect to $\omega$.
    \EndIf
    \State $j \leftarrow j-1$
    \Comment{training on the adjacent domain}
\EndWhile
}
\end{algorithmic}
\end{algorithm}

\subsection{Theoretical aspects of flow-based model}\label{sec:prop-theory}
Our proposed method maximizes the log-likelihood of labeled data with respect to $\omega$ given by Eq.~\eqref{eq:ll-label}. 
We utilize NFs to model the distribution of inputs $p(\bm{x}|y)$, and via Eq.~\eqref{eq:predict}, our proposed method also implicitly models the distribution of outputs $p(y|\bm{x})$. 
We denote the expected loss on the source domain as follows: 
\begin{align}    
    \mathrm{L_1}(g_{\omega}) 
    &\!=\! \mathbb{E}_{\bm{x},y \sim p_1(\bm{x},y)}[-\log p(y|\bm{x})] \\
    &\!=\! \mathbb{E}_{\bm{x},y \sim p_1(\bm{x},y)}\left[-\log \frac{\mathcal{N}(g_{\omega}(\bm{x},0)|\mu_y,\Sigma_y)}{\sum_{s=1}^{C} \mathcal{N}(g_{\omega}(\bm{x},0)|\mu_s,\Sigma_s)}\right].
\end{align}
Similarly, the expected loss on the $j$-th domain is denoted as $\mathrm{L_j}(g_{\omega}) \!=\! \mathbb{E}_{\bm{x},y \sim p_j(\bm{x},y)}[-\log p(y|\bm{x})]$.
For notational simplicity, we omit the argument of the expected loss and denote it as $\mathrm{L_1}$ and $\mathrm{L_j}$.
Note that $\log p(y|\bm{x})$ is a probability mass function, and we consider the following natural assumption.
\begin{assumption}[\cite{nguyen2022kl}]
    For some $M \in \mathbb{R}_{\geq 0}$, the loss satisfies $0 \leq -\log p(y|\bm{x}) \leq M$, where $\forall \bm{x} \in \mathcal{X}, \forall y \in \mathcal{Y}.$
    \label{ass:logp}
\end{assumption}

\citet{nguyen2022kl} derived an upper bound for the loss $\mathrm{L_2}$ on the basis of the source loss $\mathrm{L_1}$ and the Kullback--Leibler (KL) divergence between $p_2$ and $p_1$.  
Note that in the standard domain adaptation, there is no intermediate domain and $K=2$.
\begin{proposition}[\cite{nguyen2022kl}]
    If Assumption~\ref{ass:logp} holds, we have
        \begin{align}
            \mathrm{L_2} 
            \leq \mathrm{L_1} 
            + \frac{M}{\sqrt{2}} \sqrt{\mathrm{KL}[p_2(\bm{x})|p_1(\bm{x})] + D_{1}} 
            \label{eq:kl-bound}
        \end{align}
    where $\mathrm{KL}[\cdot|\cdot]$ denotes the KL divergence and $D_{t}=\mathbb{E}_{p_{t+1}(\bm{x})}[\mathrm{KL}[p_{t+1}(y|\bm{x})|p_t(y|\bm{x})]]$.
    \label{prop:kl-bound}
\end{proposition}

\noindent
All proofs are provided in~\ref{sec:app-proof}. 
In Eq.~\eqref{eq:kl-bound}, we note that it is impossible to calculate the conditional misalignment term $\mathbb{E}_{p_2(\bm{x})}[\mathrm{KL}[p_2(y|\bm{x})|p_1(y|\bm{x})]]$ since the labels from the second domain are not given.
We make the following covariate shift assumption
    \footnote{This assumption can be slightly relaxed by setting $\mathrm{KL}[p_{t+1}(y|\bm{x})|p_{t}(y|\bm{x})] \leq \varepsilon_t$ with small constants $\varepsilon_t \geq 0, \forall t \in \{1,\dots,K-1\}$.}:
\begin{assumption}[\cite{shimodaira2000improving}]
For any $t \in \{1,2,\dots,K\}, p_t(y|\bm{x}) = p_{t+1}(y|\bm{x})$, $\forall \bm{x} \in \mathcal{X}, \forall y \in \mathcal{Y}$. 
    \label{ass:covariate}
\end{assumption}

\noindent
We can reduce the marginal misalignment term $\mathrm{KL}[p_2(\bm{x})|p_1(\bm{x})]$ by transforming $p_2$ to $p_1$ with NFs.
A CNF transforms a sample $\bm{x} \sim p_{t+1}(\bm{x})$ to a sample from the probability distribution $p_t(\bm{x})$ of the adjacent domain, and the likelihood of the model satisfies 
\begin{equation}
    \log p_{t+1}(g_{\omega}(\bm{x},t+1)) 
    \!=\! \log p_{t}(g_{\omega}(\bm{x},t)) 
    - \int_{t}^{t+1} \mathrm{Tr} \left(\frac{\partial v}{\partial g_{\omega}}\right) dt.
    \label{eq:cnf-partial}
\end{equation}
The expectation of the minus of the log-likelihood function to be minimized is given by
\begin{equation}
    \mathbb{E}_{p_{t+1}(\bm{x})}[-\log p_{t+1}(g_{\omega}(\bm{x},t+1))].
    \label{eq:cnf-expected}
\end{equation}

\noindent
\citet{onken2021ot} derived the following proposition.
\begin{proposition}[\cite{onken2021ot}]
    The minimization of Eq.~\eqref{eq:cnf-expected} is equivalent to the minimization of the KL divergence between $p_{t}(\bm{x})$ and $p_{t+1}(\bm{x})$ transformed by $g_{\omega}$.
    \label{prop:cnf}
\end{proposition}

\noindent
Let $p_{t+1}^{\ast}(\bm{x})$ be the flowed distribution obtained by transforming $p_{t+1}$ using $g$ from time $t+1$ to $t$.
Proposition~\ref{prop:cnf} allows us to rewrite Eq.~\eqref{eq:kl-bound} as follows:
\begin{align}
    \mathrm{L_2} 
    \leq \mathrm{L_1} + \frac{M}{\sqrt{2}} \sqrt{\mathrm{KL}[p_2^{\ast}(\bm{x})|p_1(\bm{x})] + D_1}.
    \label{eq:kl-bound-logp}
\end{align}
We extend Eq.~\eqref{eq:kl-bound-logp} and introduce the following corollary that gives an upper bound of the target loss $\mathrm{L_K}$.
\begin{corollary}
If Assumptions~\ref{ass:logp} and~\ref{ass:covariate} hold, we have
\begin{equation}
    \mathrm{L_K} 
    \leq \mathrm{L_1} + \frac{M}{\sqrt{2}} \sum_{t=2}^{K} \sqrt{\mathrm{KL}[p_t^{\ast}(\bm{x})|p_{t-1}(\bm{x})]}.
\end{equation}
\label{corollary}
\end{corollary}

In gradual domain adaptation, it is assumed that there is a large discrepancy between the source and target domains; hence, it is highly likely that $p_{K}(\bm{x})/p_{1}(\bm{x}) \to \infty$ for some $\bm{x}$ and the KL divergence between the marginal distributions of the source and the target is not well-defined.  
Corollary~\ref{corollary} suggests that our proposed method avoids this risk by utilizing the intermediate domains and that the target loss $\mathrm{L_K}$ is bounded. 
In Section~\ref{sec:sub-exp-nointer}, we show that the target loss $\mathrm{L_K}$ is large when our flow-based model $g_{\omega}$ is trained only with the source and target datasets. 

Compared to the upper bound for the self-training-based GDA Eq.~\eqref{eq:bounds}, which depends only on the loss in the source domain and the number of the intermediate domains, our bound is adaptive that incorporates trained CNF in the KL-divergence term. 
The self-training procedure contains hyperparameters such as the number of epochs and learning rates, and how to determine those appropriate hyperparameters based on Eq.~\eqref{eq:bounds} is non-trivial.
In contrast, Corollary~\ref{corollary} is free from self-training.

\subsection{Scalability}\label{sec:prop-limit}
\citet{grathwohl2018ffjord} discussed the details of the scalability of a CNF.
They assumed that the cost of evaluating a CNF $g$ is $\mathcal{O}(dH)$, where $d$ is the dimension of the input and $H$ is the size of the largest hidden unit in $v$. 
They derived the cost of computing the likelihood as $\mathcal{O}(dHN)$, where $N$ is the number of evaluations of $g$ in the ODE solver. 
In general, the training cost of CNF $g$ is high because the number of evaluations $N$ of $g$ in the ODE solver is large. 
When $K$ domains are given, we have to solve the initial value problem $\frac{1}{2} K(K+1)$ times in our proposed method, resulting the computational cost $\mathcal{O}(dHNK^2)$, which is computationally expensive when the number of intermediate domains is large. 
When we can access many intermediate domains and the distances between the intermediate domains are small, the conventional self-training-based GDA algorithm~\citep{kumar2020understanding} will be suitable.

\section{Experiments}\label{sec:exp} 
We use the implementation of the CNF\footnote{\url{https://github.com/hanhsienhuang/CNF-TPR}} provided by~\citet{huang2021accelerating}. 
PyNNDescent\footnote{\url{https://github.com/lmcinnes/pynndescent/tree/master}} provides a python implementation of the nearest neighbor descent~\citep{dong2011efficient}. 
PyTorch~\citep{NEURIPS2019_9015} is used to implement all procedures in our proposed method except for the procedure of CNF and nearest neighbor descent. 
The details of the experiment, such as the composition of the neural network, are presented in~\ref{sec:app-exp-detail}.
We use WILDS~\citep{koh2021wilds} and MoleculeNet~\citep{wu2018moleculenet} to load pre-processed datasets. 
All experiments are conducted on our server with Intel Xeon Gold 6354 processors and NVIDIA A100 GPU. 
Source code to reproduce the experimental results is available at \url{https://github.com/ISMHinoLab/gda_via_cnf}.

\begin{table}[tbp]
\caption{Summary of datasets. 
{\tt Rotating MNIST}, {\tt Portraits}, and {\tt RxRx1} are image datasets. 
{\tt Two Moon} and {\tt Block} are toy datasets.
{\tt Block} has two intermediate domains.
}
\begin{center}
\scalebox{1}{
\begin{tabular}{lccccc}
    \hline
    \multicolumn{1}{c}{\multirow{2}{*}{Name}} & 
    \multicolumn{1}{c}{\# of} & 
    \multicolumn{1}{c}{\# of} & 
    \multicolumn{3}{c}{\# of Samples} \\
    \cline{4-6}
    &  
    Dimensions &  
    Classes & 
    Source & 
    Intermediate & 
    Target \\
    \hline
    {\tt Two Moon} & 
    2 &
    2 &
    1,788 &
    1,833 &
    1,818 \\
    {\tt Block} &
    2 &
    5 &
    1,810 &
    1,840/1,845 &
    1,840\\
    {\tt Rotating MNIST} &
    \multirow{2}{*}{$28 \!\times\! 28$} &
    \multirow{2}{*}{10} &
    \multirow{2}{*}{2,000} &
    \multirow{2}{*}{2,000} &
    \multirow{2}{*}{2,000} \\
    \citep{kumar2020understanding} \\
    {\tt Portraits}&
    \multirow{2}{*}{$32 \!\times\! 32$} &
    \multirow{2}{*}{2} &
    \multirow{2}{*}{2,000} &
    \multirow{2}{*}{2,000} &
    \multirow{2}{*}{2,000} \\
    \citep{ginosar2015century} \\
    {\tt SHIFT15M}&
    \multirow{2}{*}{4,096} &
    \multirow{2}{*}{7} &
    \multirow{2}{*}{5,000} &
    \multirow{2}{*}{5,000} &
    \multirow{2}{*}{5,000} \\
    \citep{kimura2021shift15m} \\
    {\tt RxRx1}&
    \multirow{2}{*}{$3 \!\times\! 32 \!\times\! 32$} &
    \multirow{2}{*}{4} &
    \multirow{2}{*}{9,856} &
    \multirow{2}{*}{9,856} &
    \multirow{2}{*}{6,856}\\
    \citep{taylor2019rxrx1} \\
    {\tt Tox21 NHOHCount}&
    \multirow{2}{*}{108} &
    \multirow{2}{*}{2} &
    \multirow{2}{*}{3,284} &
    \multirow{2}{*}{1,898} &
    \multirow{2}{*}{807} \\
    \citep{thomas2018us} \\
    {\tt Tox21 RingCount}&
    \multirow{2}{*}{108} &
    \multirow{2}{*}{2} &
    \multirow{2}{*}{1,781} &
    \multirow{2}{*}{2,308} &
    \multirow{2}{*}{1,131} \\
    \citep{thomas2018us} \\
    {\tt Tox21 NumHDonors}&
    \multirow{2}{*}{108} &
    \multirow{2}{*}{2} &
    \multirow{2}{*}{3,246} &
    \multirow{2}{*}{2,285} &
    \multirow{2}{*}{835} \\
    \citep{thomas2018us} \\
    \hline
\end{tabular}
}
\end{center}
\label{tab:dataset}
\end{table}

\subsection{Datasets}\label{sec:exp-data}
We use benchmark datasets with modifications for GDA. 
Since we are considering the situation that the distances of the adjacent domains are large, we prepare only one or two intermediate domains. 
We summarize the information on the datasets used in our experiments in Table~\ref{tab:dataset}. The details of the datasets are shown in~\ref{sec:app-exp-detail}.

\subsection{Experimental settings}\label{sec:exp-setup}
\citet{brehmer2020flows} mentioned that NFs are unsuitable for data that do not populate the entire ambient space. 
Therefore, we apply UMAP~\citep{mcinnes2018umap-software} to each dataset as preprocessing. 
UMAP can use partial labels for semi-supervised learning; hence, we utilize the labels of the source dataset as the partial labels. 
To determine the appropriate embedding dimension, we train a CNF $g_{\omega}$ on the dimension-reduced source dataset by maximizing Eq.~\eqref{eq:ll-label} with respect to $\omega$. 
After the training, we evaluate the accuracy on the dimension-reduced source dataset. 
We select the embedding dimension with which the result of mean accuracy of three-fold cross-validation in the dimension-reduced source domain is the best. 
All parameters in UMAP are set to their default values.

\subsection{Learning gradual shifts with discrete normalizing flows}\label{sec:sub-exp-dnf} 
We aim to learn the continuous change $\bm{x}^{(K)} \mapsto \cdots \mapsto \bm{x}^{(1)} \mapsto \bm{z}$ by using NFs. 
In principle, we can map the distribution of the target domain to the Gaussian mixture distribution via the intermediate and the source domains by utilizing multiple discrete NFs $f^{(K)}, \ldots, f^{(1)}$. However, it is empirically shown that DNFs are unsuitable for learning continuous change. 

\begin{figure*}[!htbp]
\centering
  \includegraphics[clip, width=12cm]{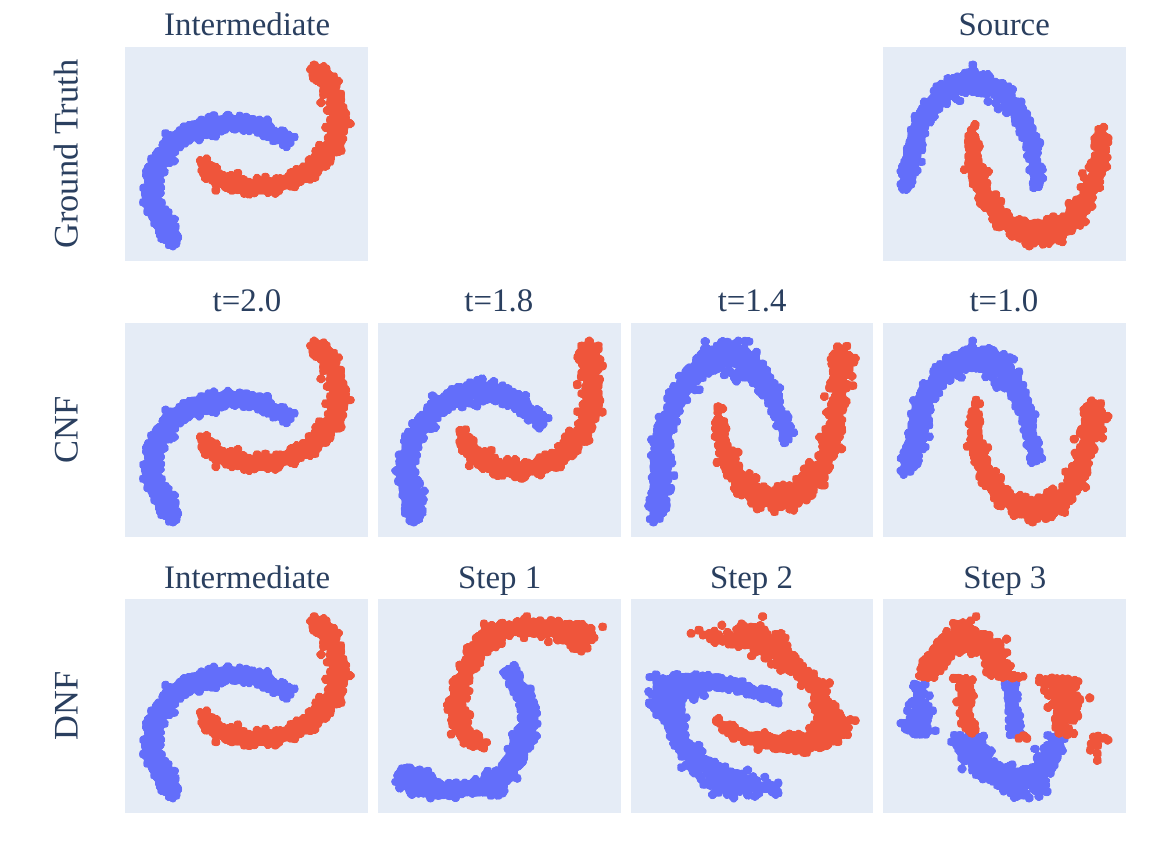}
  \caption{
    Comparison between the discrete and the continuous normalizing flows.
    Whereas continuous NFs are suitable for learning continuous change, discrete NFs are unsuitable for learning continuous change.
  }
  \label{fig:realnvp}
\end{figure*}

We use RealNVP~\citep{DBLP:conf/iclr/DinhSB17} as a DNF. 
One DNF block consists of four fully connected layers with 64 nodes in each layer. 
To improve the expressive power of the flow-based model, we stack three DNF blocks. 
The output from each DNF block should vary continuously. 
We train the DNFs with the source and the intermediate datasets on the {\tt Two Moon} dataset. 
Figure~\ref{fig:realnvp} shows the transformation of the intermediate data to the source data. 
We see that the CNF continuously transforms the intermediate data into the source data. 
On the other hand, the DNFs failed to transform the intermediate data into the source data, and the path of transformation is not continuous. 
From this preliminary experiment, we adopt continuous normalizing flow to realize the proposed gradual domain adaptation method. 
As a byproduct, the proposed method can generate synthetic data from any intermediate domain even when no sample from the domain is observed. 

\subsection{Estimation of log-likelihood by fitting Gaussian mixture distribution}\label{sec:sub-exp-fitted_gmm}
Our proposed method learns the transformation of a sample from the $j$-th domain to a sample from the adjacent domain, and it requires the estimation of the log-likelihood $\log p_{t-1}(g_{\omega}(\bm{x}^{(t)}, t\!-\!1))$, where $t=j$. 
We proposed a computation method for the log-likelihood by using $k$NN estimators in Section~\ref{sec:prop-knn}. 
Here, as another way of estimating the log-likelihood, we consider the approximation of $p_{t-1}$ by a Gaussian mixture distribution.
Let $Q$ and $w_{q}^{(t-1)}$ be the number of mixture components and a mixture weight, respectively. 
The subscript $q$ indicates the $q$-th component of a Gaussian mixture distribution, and the superscript indicates the domain. 
We approximate the adjacent domain with the Gaussian mixture distribution
\begin{equation}
    p_{t-1}(\bm{x}^{(t-1)})
    =\sum_{q=1}^{Q} w_{q}^{(t-1)} \mathcal{N}(\bm{x}^{(t-1)}|\bm{\mu}_{q}^{(t-1)}, \Sigma_{q}^{(t-1)}),
    \;
    \sum_{q=1}^{Q} w_{q}^{(t-1)} = 1,
\end{equation}
where $\bm{\mu}_{q}^{(t-1)}$ and $\Sigma_{q}^{(t-1)}$ are the mean vector and the covariance matrix, respectively. 
We fit a Gaussian mixture distribution for each domain. 
Therefore, we distinguish the mixture weight, the mean vector, and the covariance matrix with superscripts. 
We assign a sufficiently large value to $Q$ since our aim is an estimation of the log-likelihood $\log p_{t\!-\!1}(g(\bm{x}^{(t)},t\!-\!1))$.

\begin{figure*}[!htbp]
\centering
  \includegraphics[clip, width=12cm]{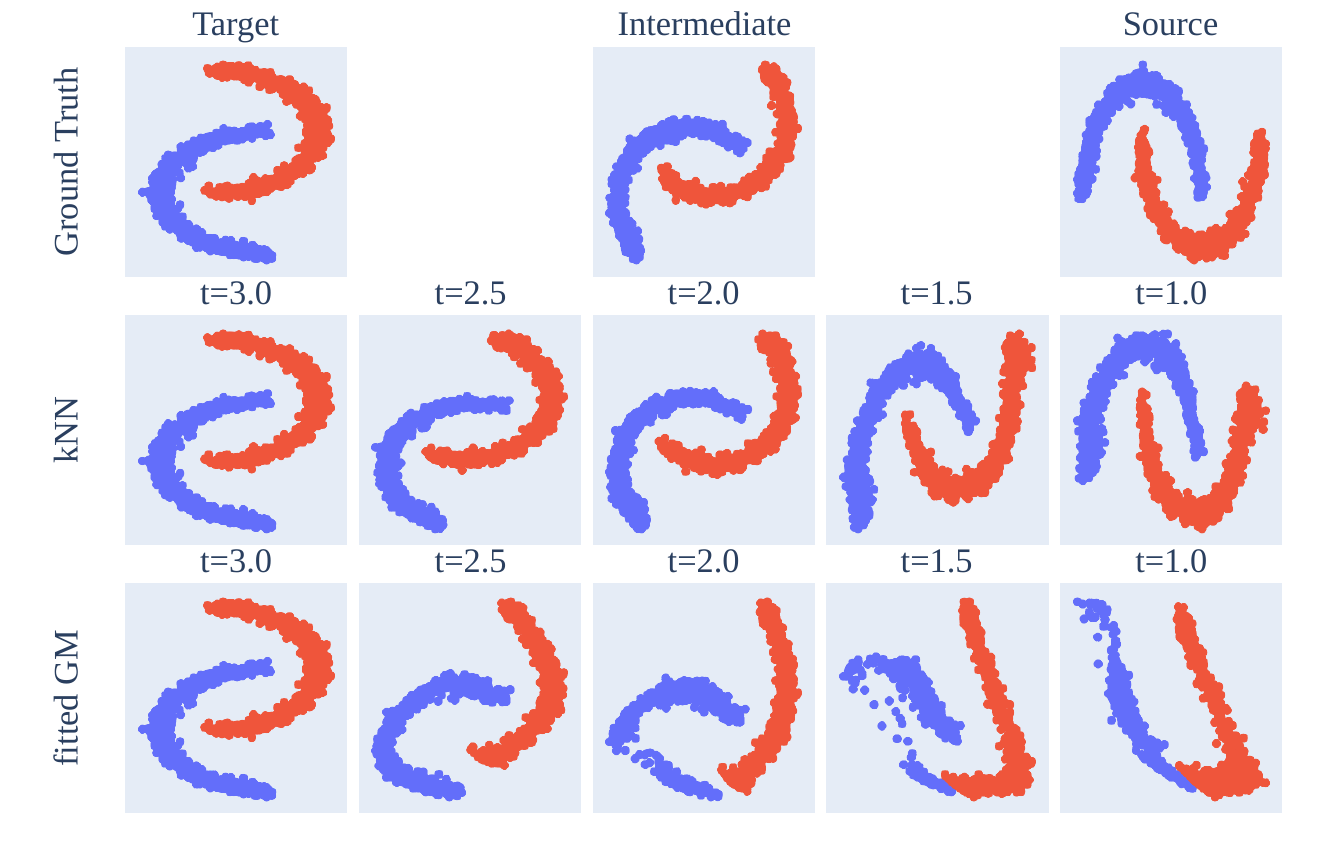}
  \caption{
  Comparison of the methods of estimating $\log p_{t\!-\!1}(g(\bm{x}^{(t)},t\!-\!1))$. 
  The CNF trained with $k$NN estimators transforms the target data to the source data as expected
  }
  \label{fig:fitted_gmm}
\end{figure*}

We compare the log-likelihood estimation by fitted Gaussian mixture distributions with that by $k$NN estimators on the {\tt Two Moon} dataset, and conclude that the $k$NN estimators are suitable for our proposed method.
Since our toy dataset is a simple two moon forms, $Q=30$ should be enough for modeling its distribution with high precision. 
Figure~\ref{fig:fitted_gmm} shows the comparison of the transformation by the CNF trained with $k$NN estimators and that trained with the Gaussian mixture distributions for evaluating the likelihood. 
Whereas the CNF trained with $k$NN estimators transforms the target data to the source data as expected, the CNF trained with fitted Gaussian mixture distributions fails to do so.

\subsection{Hyperparameters}\label{sec:sub-exp-param} 
The proposed method has two hyperparameters, $k$ and $r$. 
These hyperparameters are introduced in Sections~\ref{sec:prop-knn} and~\ref{sec:prop-gm}, respectively.
The hyperparameter $k$ affects the computation of log-likelihood by $k$NN estimators, and the hyperparameter $r$ controls the distance between the Gaussian distributions corresponding to each class. 
In this section, from a practical view point, we discuss how to tune these hyperparameters. 

First, we discuss the tuning method of $k$. 
We estimate the log-likelihood by using a $k$NN estimator when learning the transformation between the consecutive domains. 
In general, the parameter $k$ controls the trade-off between bias and variance. 
Optimal $k$ also depends on whether the given dataset has local fine structures or not. 
We determine an appropriate $k$ by fitting a $k$NN classifier on the source dataset. 
We train the $k$NN classifier with only the source dataset and determine $k$ from the result of three-fold cross-validation.

\begin{figure*}[!htbp]
\centering
  \includegraphics[clip, width=14cm]{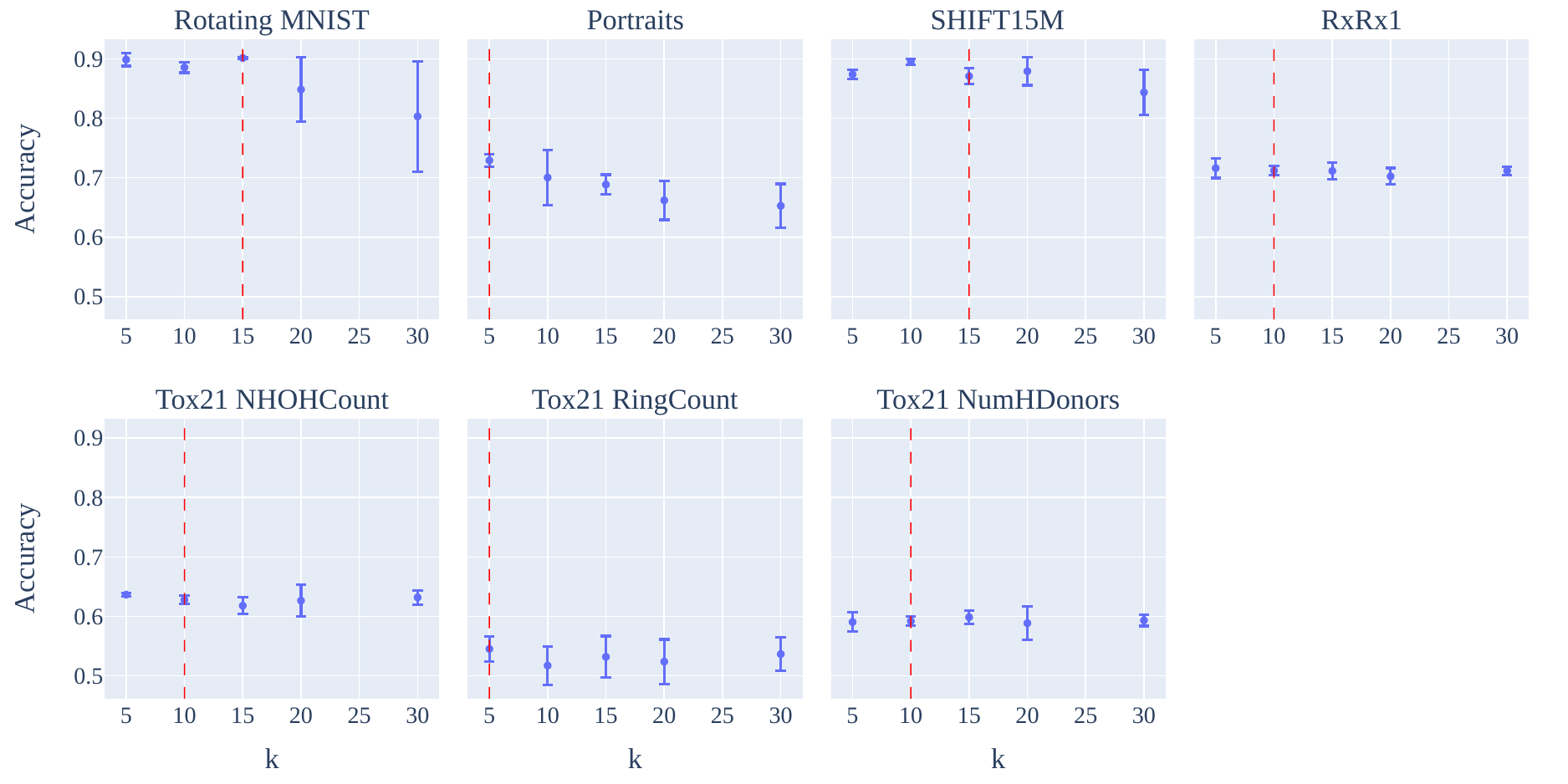}
  \caption{
  Experimental result of the training of our flow-based model with various hyperparameter $k$ values. 
  The appropriate $k$ can be determined roughly by the fitting of the $k$NN classifier on the source dataset.
  }
  \label{fig:knn}
\end{figure*}

We vary the hyperparameter $k$ and train our flow-based model $g_{\omega}$. 
After the training, we evaluate the accuracy on the target dataset. 
The evaluation of our flow-based model was repeated three times using different initial weights of neural networks. 
In Figure~\ref{fig:knn}, the red dashed line denotes the $k$ determined using the result of $k$NN classifier fitting. 
The appropriate $k$ can be determined roughly by the fitting of the $k$NN classifier on the source dataset. 

Next, we discuss a tuning method for $r$. 
To conduct gradual domain adaptation with an NF, our flow-based model learns the transformation between the distribution of the source domain and a Gaussian mixture distribution. 
The Gaussian distributions for different labels should be distinguishable from each other.
We assign the mean of each Gaussian distribution with the polar coordinates system. 
Therefore, we should set the appropriate $r$, which is the distance from the origin in the polar coordinate system, on the basis of the number of classes.

\begin{figure}[htbp]
\centering
  \begin{minipage}[t]{0.45\linewidth}
    \centering
    \includegraphics[keepaspectratio, scale=0.45]{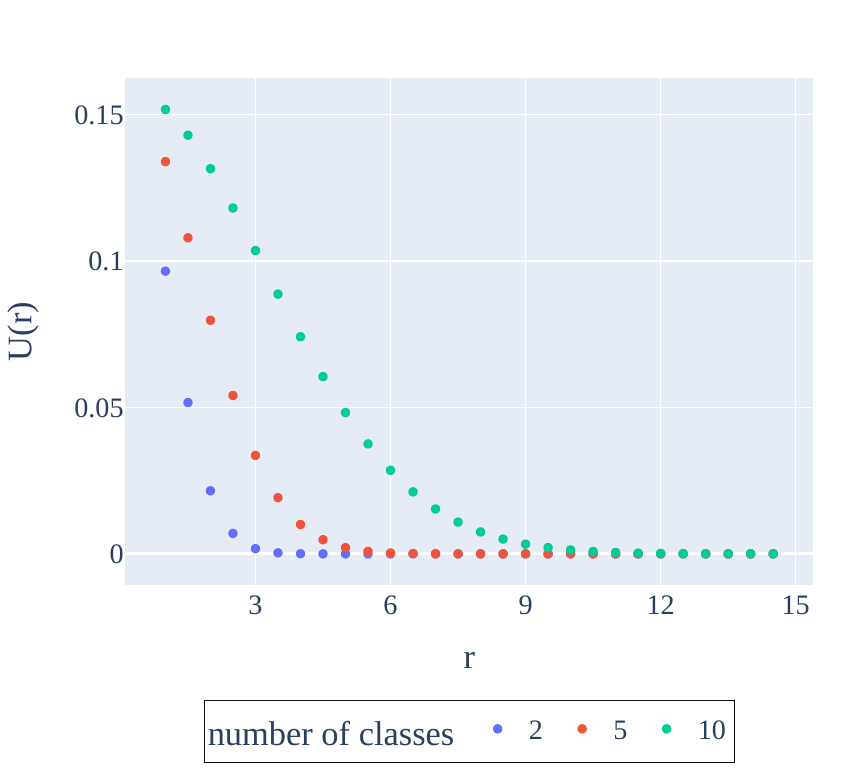}
    \subcaption{Number of dimensions is fixed.}
    \label{fig:mean_r_calc_a}
  \end{minipage}
  \begin{minipage}[t]{0.45\linewidth}
    \centering
    \includegraphics[keepaspectratio, scale=0.45]{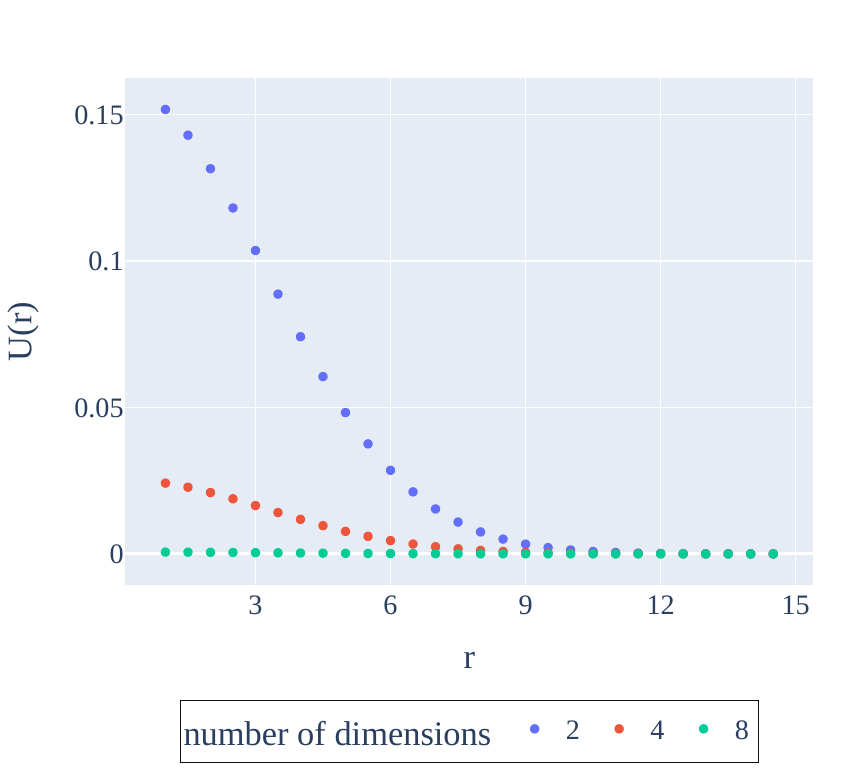}
    \subcaption{Number of classes is fixed.}
    \label{fig:mean_r_calc_b}
  \end{minipage}
  \caption{  
  Determination method of hyperparameter $r$.
  We can roughly determine the hyperparamter $r$ by considering the number of classes and the number of dimensions of data.
  }
  \label{fig:mean_r_calc}
\end{figure}

The hyperparameter $r$ can be roughly determined by considering the number of classes and the number of dimensions of data. 
The distribution $p_0$, conditioned on the label $s$, follows the Gaussian distribution $\mathcal{N}(\bm{\mu}_s, \Sigma_s)$.
Let $\bm{m}(r)$ be the midpoint vector of the mean vectors $\bm{\mu}_s$ and $\bm{\mu}_s'$ of two adjacent Gaussian distributions. We assign a mean vector with the polar coordinates system as shown in Eq.~\eqref{eq:mean_r}, and it depends on the hyperparameter $r$.
We propose a method to determine $r$ by calculating $\max(\mathcal{N}(\bm{m}(r)|\bm{\mu}_s, \Sigma_s), \mathcal{N}(\bm{m}(r)|\bm{\mu}_s', \Sigma_s'))$, and for simplicity, we denote it as $U(r)$. 
We should select a sufficiently small $r$ such that $U(r) \simeq 0$ since the Gaussian distributions for different labels should be separable.
We vary the hyperparameter $r$ and calculate $U(r)$. 
Figure~\ref{fig:mean_r_calc} shows the result of the calculation. 
In Figure~\ref{fig:mean_r_calc}~\subref{fig:mean_r_calc_a}, when the number of dimensions is fixed at 2, it can be seen that a larger $r$ is required with a large number of classes. 
In Figure~\ref{fig:mean_r_calc}~\subref{fig:mean_r_calc_b}, the number of classes is fixed at 10, and we see that $U(r)$ becomes sufficiently small even for a small $r$ when the number of dimensions is large.

\begin{figure*}[!htbp]
\centering
  \includegraphics[clip, width=14cm]{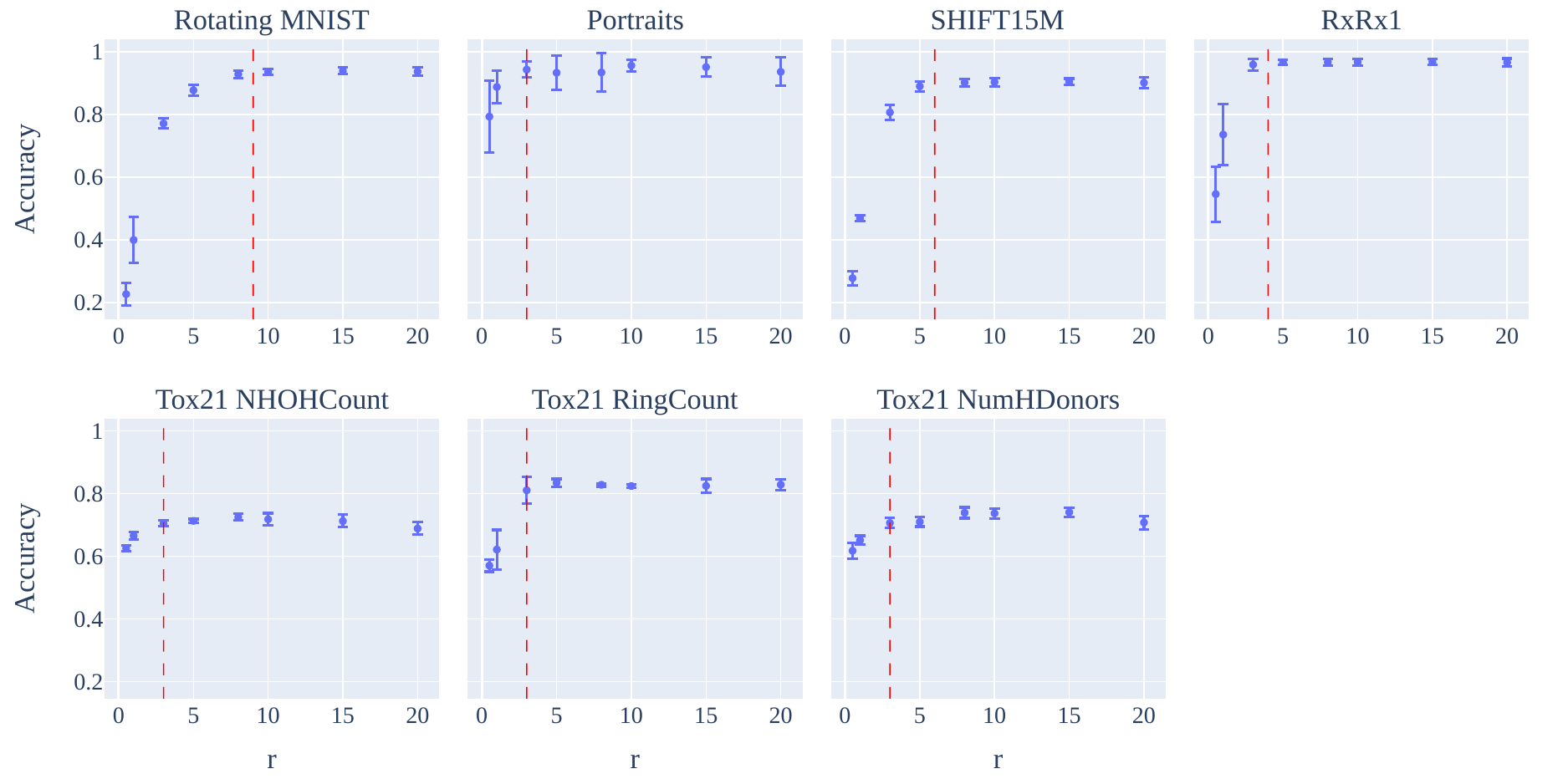}
  \caption{
  Experimental results of the training of our flow-based model with various hyperparameter $r$ values on the source dataset only. 
  When the hyperparameter $r$ is sufficiently large, the accuracy on the source dataset does not change significantly.
  }
  \label{fig:mean_r}
\end{figure*}

We vary the hyperparameter $r$ and train the proposed flow-based model $g_{\omega}$ with only the source dataset, i.e., we maximize the log-likelihood given by Eq.~\eqref{eq:ll-label} with respect to $\omega$. 
The performance of our flow-based model is evaluated using three-fold cross-validation on the source dataset.
Figure~\ref{fig:mean_r} shows the result of mean accuracy of three-fold cross-validation. 
The red dashed line in Figure~\ref{fig:mean_r} represents the smallest $r$, which induces $U(r) < 0.001$. 
We see that the accuracy on the source dataset does not change significantly with a sufficiently large $r$, which means that the Gaussian distributions for different labels are distinguishable from each other. 
Therefore, we should determine the hyperparameter $r$ by calculating $U(r)$.

\subsection{Necessity of intermediate domains}\label{sec:sub-exp-nointer} 
\begin{figure}[!htbp]
\centering
  \includegraphics[clip, width=12cm]{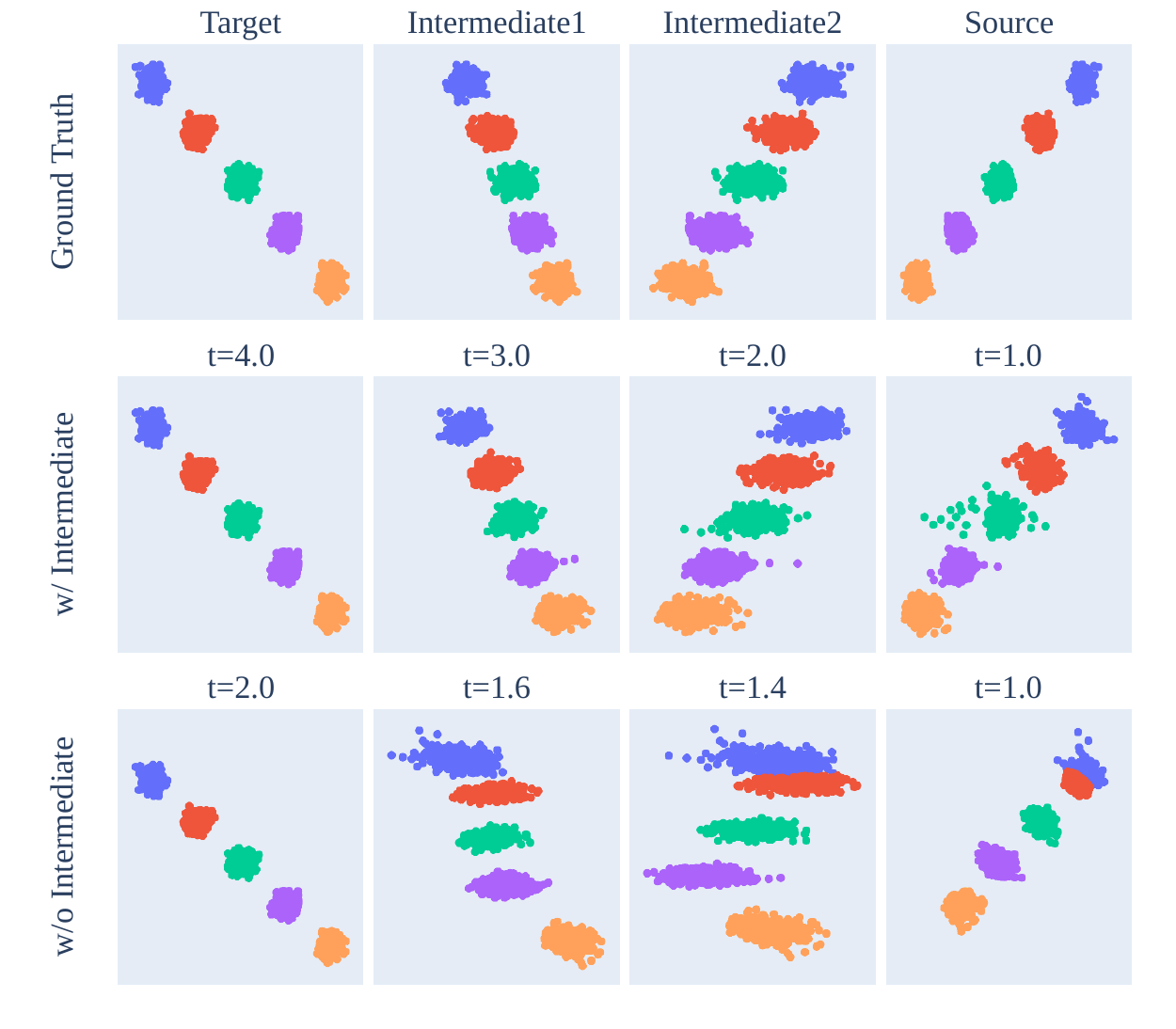}
  \caption{
    Necessity of intermediate domains.
    The CNF trained with the intermediate dataset transforms the target data to the source data as expected. 
    From the perspective of predictive performance on the target data, training CNF with the intermediate datasets is preferable. 
  }
  \label{fig:no_inter}
\end{figure}

Our key idea is to use a CNF to learn gradual shifts between domains. 
We show the necessity of the intermediate domains for the training of the CNF. 
Figure~\ref{fig:no_inter} shows the results of CNF trained with and without the intermediate datasets on the {\tt Block} dataset. 
In Figure~\ref{fig:no_inter}, we show the transformation from the target data to the source data by the trained CNF. 
Visually, the CNF trained with the intermediate dataset transforms the target data to the source data as expected. 
The accuracy on the target dataset is 0.999 when CNF is trained with the intermediate datasets.
In contrast, the accuracy on the target dataset is 0.181 when CNF is trained without intermediate datasets.
From these results, we conclude that it is important to train a CNF with datasets from the source, intermediate, and target domains to capture the gradual shift. 

\citet{pmlr-v162-wang22n} mentioned that the sequence of intermediate domains should be uniformly placed along the Wasserstein geodesic between the source and target domains.
When only one intermediate domain is given, intuitively, a preferable intermediate domain lies near the midpoint of the Wasserstein geodesic between the source and target domains.
In practice, we cannot select arbitrary intermediate domains for domain adaptation, and the sequence of intermediate domains is only given.

\begin{figure}[htbp]
\centering
  \begin{minipage}[t]{0.6\columnwidth}
    \centering
    \includegraphics[keepaspectratio, scale=0.45]{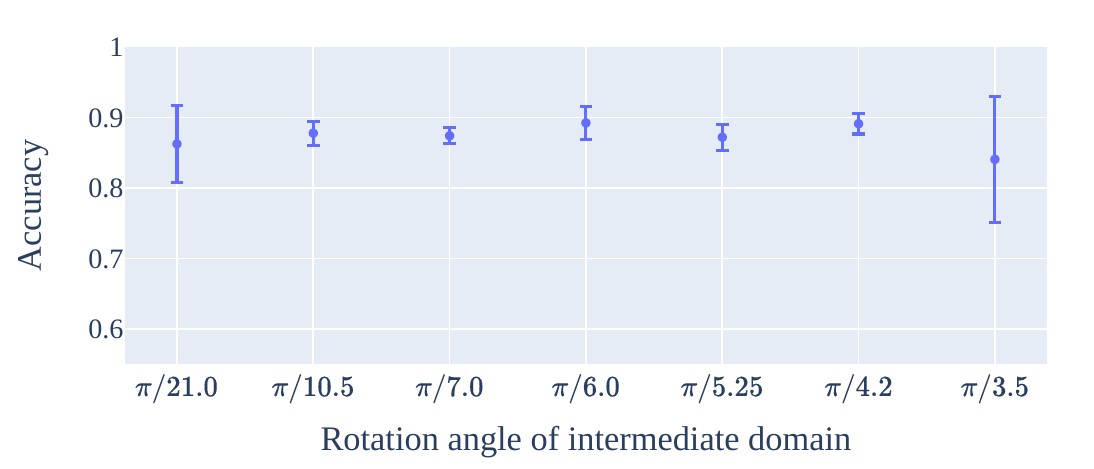}
    \subcaption{Various intermediate domains.}
    \label{fig:mnist_inter_a}
  \end{minipage}
  \begin{minipage}[t]{0.3\columnwidth}
    \centering
    \includegraphics[keepaspectratio, scale=0.45]{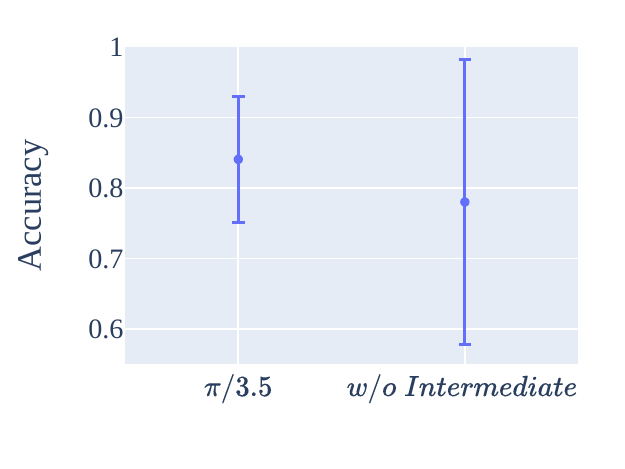}
    \subcaption{Necessity of intermediate domains.}
    \label{fig:mnist_inter_b}
  \end{minipage}
  \caption{  
  Experimental results of the training of our flow-based model with various intermediate domains. 
  }
  \label{fig:mnist_inter}
\end{figure}

In the {\tt Rotating MNIST} dataset, the source dataset consists of samples from the {\tt MNIST} dataset without any rotation, and the target dataset is prepared by rotating images of the source dataset by angle $\pi/3$.
We consider a preferable intermediate domain to be a dataset prepared by rotating images from the source dataset by angle $\pi/6$.
We vary the rotation angle when preparing the intermediate domain and train our flow-based model.
Note that the number of intermediate domains used for training is always one.
After the training, we evaluate the accuracy on the target dataset. 
The evaluation of our flow-based model was repeated five times using different initial weights of neural networks.
In Figure~\ref{fig:mnist_inter}~\subref{fig:mnist_inter_a}, we see that rotations with small and large angles, i.e., $\pi/21$ and $\pi/3.5$, show a significant variance in accuracy, and the mean accuracy for these angles is lower than that for $\pi/6$.
We obtain the experimental results that support our intuition, indicating that the preferable rotation angle for the intermediate domain is $\pi/6$.
Figure~\ref{fig:mnist_inter}~\subref{fig:mnist_inter_b} shows the results of comparison between the worst-performing model from Figure~\ref{fig:mnist_inter}~\subref{fig:mnist_inter_a} and the model trained without any intermediate domain.
The model trained without intermediate domains performs the worst. 
The quality of the intermediate domain does impact the results of gradual domain adaptation, but the impact is relatively small compared to the result obtained without the intermediate domain.

\subsection{Generating artificial intermediate domains}\label{sec:sub-exp-generate}
While our primary focus lies in GDA, our proposed method can generate synthetic data from intermediate domains, even in the absence of observed samples from that domain.

\begin{figure}[!htbp]
\centering
  \includegraphics[clip, width=12cm]{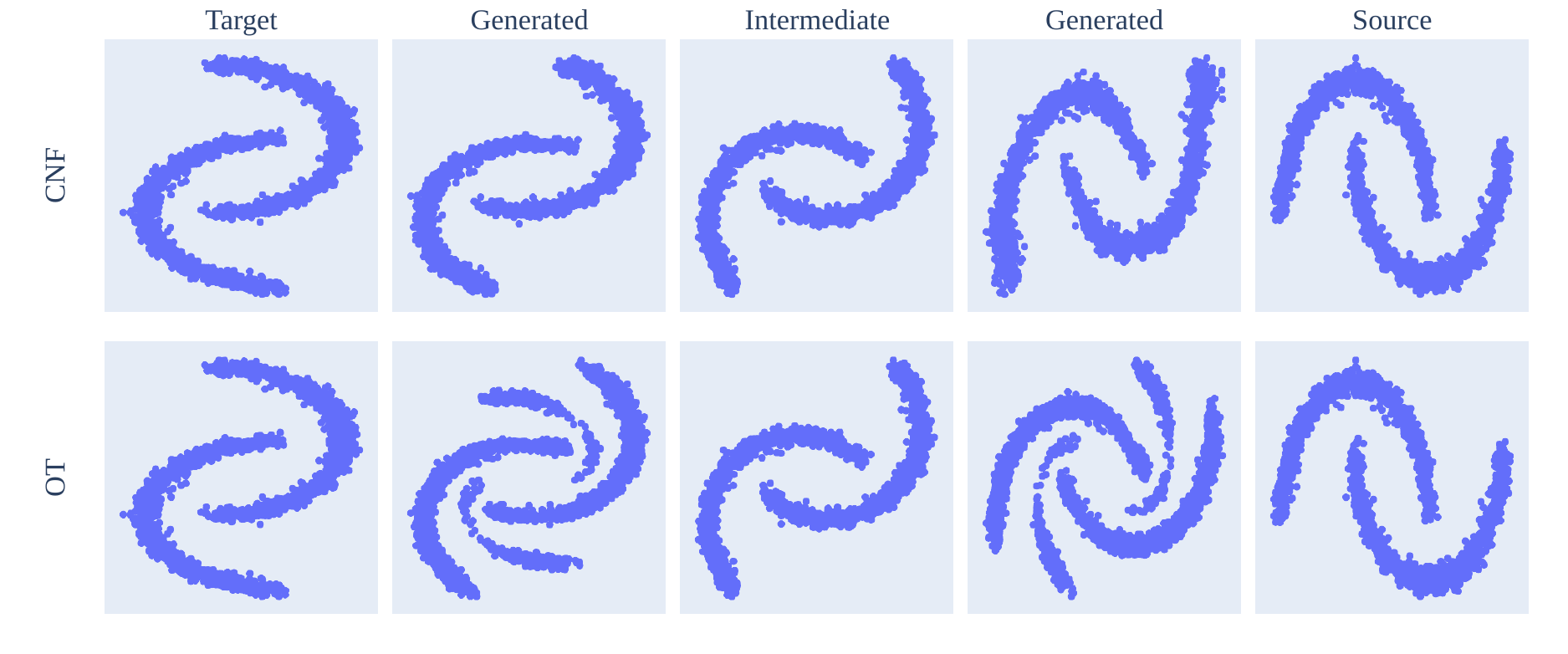}
  \caption{
  Comparison of the methods of generating pseudo-intermediate domains on the {\tt Two Moon} dataset.
  CNF generates reasonable pseudo-intermediate domains, whereas OT fails to do so.
  }
  \label{fig:ot}
\end{figure}

We aim to utilize NFs to transform the distribution of the target domain to the distribution of the source domain. 
Optimal transport (OT) can also realize a natural transformation from the target domain to the source domain.
\citet{he2023gradual} proposed a method called \textbf{G}enerative Gradual D\textbf{O}main \textbf{A}daptation with Optimal \textbf{T}ransport (GOAT).
GOAT interpolate the initially given domains with OT and apply gradual self-training.
It is important to obtain appropriate pseudo-intermediate domains using OT since GOAT is a self-training-based GDA algorithm.
Figure~\ref{fig:ot} shows the comparison results between pseudo-intermediate domains generated by the proposed method and those generated by OT.
On the {\tt Two Moon} dataset, CNF is suitable for generating pseudo-intermediate domains, whereas OT is unsuitable for generating pseudo-intermediate domains. 

\begin{figure}[tbp]
\centering
  \includegraphics[clip, width=11cm]{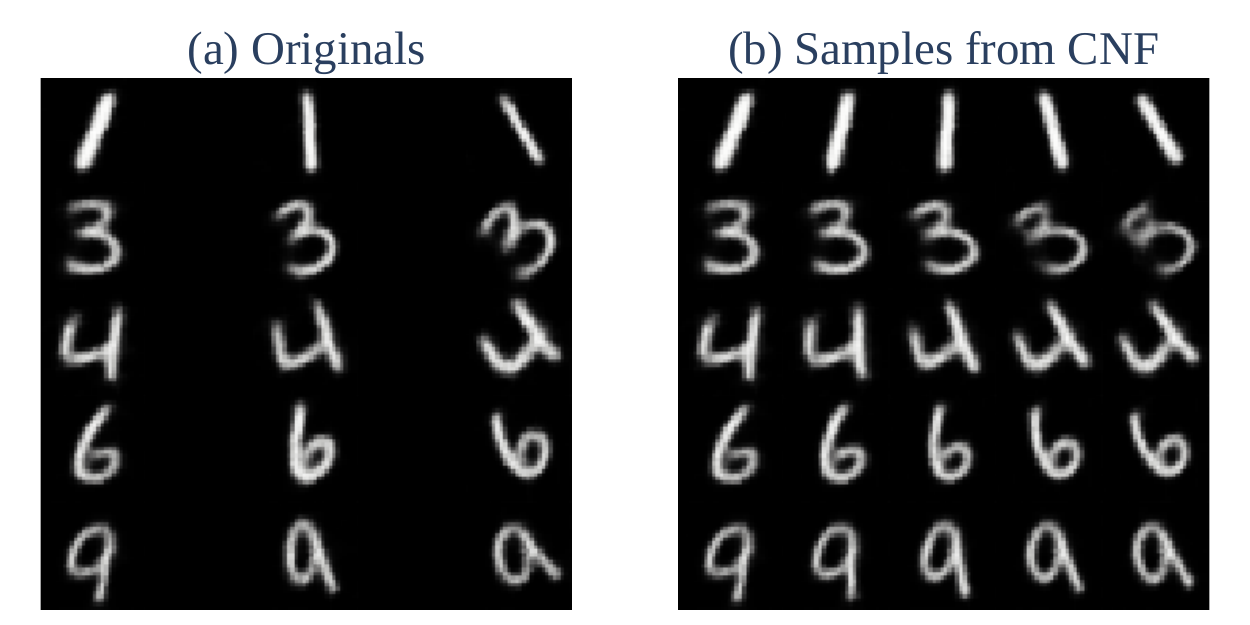}
  \caption{Morphing of {\tt Rotating MNIST}.}\label{fig:vae}
\end{figure}

In principle, the proposed method is applicable to any dimensional data as input, such as image data, but the current technology of CNF is the computational bottleneck. 
It is difficult to handle high-dimensional data directly due to the high-computational cost of off-the-shelf CNF implementation. 
Moreover, as mentioned by~\citet{brehmer2020flows}, NFs are unsuitable for data that do not populate the entire ambient space. 
Therefore, as a preprocessing, it is reasonable to reduce the dimensionality of high-dimensional data. 
A combination of the proposed method and VAE is applicable to image data. 
We can utilize the trained CNF and VAE for generating artificial intermediate images such as morphing. 
We show a demonstration of morphing on {\tt Rotating MNIST} in Figure~\ref{fig:vae}. 
The details of the experiment are shown in~\ref{sec:app-exp-detail}.

\subsection{Comparison with baseline methods}\label{sec:sub-exp-compare}
To verify the effectiveness of the proposed method, we compare it with the baseline methods. 
Following the approach described in Section~\ref{sec:sub-exp-param}, we assign different values to the hyperparameter $k$ of the proposed method for each dataset. 
The hyperparameter $r$ is set to $r=3$ and $r=10$ for binary and multiclass classification, respectively. 
The appropriateness of these settings is discussed in Section~\ref{sec:sub-exp-param}.
The primary baseline methods are self-training-based GDA methods, as introduced in Section~\ref{sec:gda}. 
Recall that these methods update the hypothesis $h: \mathcal{X} \to \mathcal{Y}$ trained on the source dataset by applying sequential self-training. 
The key idea of GDA is that the hypothesis $h$ should be updated gradually. 
GIFT~\citep{abnar2021gradual} and AuxSelfTrain~\citep{zhang2021gradual} are methods that apply the idea of GDA to conventional domain adaptation. 
While conventional (non-gradual) domain adaptation is beyond the scope of this study, we limit our comparisons to those methods inspired by GDA.
Although GIFT and AuxSelfTrain do not utilize intermediate domains, we also show experimental results using intermediate domains for updating the hypothesis sequentially (Sequential GIFT and Sequential AuxSelfTrain).
EAML~\citep{NEURIPS2020_fd69dbe2} is a method that uses meta-learning to adapt to a target domain that evolves over time.
We compare the proposed method with EAML because the problem setting assumed by EAML is very similar to that assumed in GDA.
The details of the experiment, such as the composition of the neural network, are described in~\ref{sec:app-exp-detail}
We provide a brief description of the baseline methods as follows.
\setlength{\leftmargini}{12pt} 
\begin{itemize}
    \setlength{\parskip}{0pt}
    \setlength{\itemsep}{0pt}
    \setlength{\labelsep}{0pt}
    \setlength{\parsep}{0pt}
    \item SourceOnly: Train the classifier with the source dataset only.
    \item GradualSelfTrain~\citep{kumar2020understanding}: Apply gradual self-training with the initially given domains.
    \item GOAT~\citep{he2023gradual}: Interpolate the initially given domains with optimal transport and apply gradual self-training.
    \item GIFT~\citep{abnar2021gradual}: Update the source model by gradual self-training with pseudo intermediate domains generated by the source and target domains.
    \item Sequential GIFT~\citep{abnar2021gradual}: Apply the GIFT algorithm with the initially given domains. 
    \item AuxSelfTrain~\citep{zhang2021gradual}: Update the source model by gradual self-training with pseudo intermediate domains generated by the source and target domains. 
    \item Sequential AuxSelfTrain~\citep{zhang2021gradual}: Apply the AuxSelfTrain algorithm with the initially given domains.
    \item EAML~\citep{NEURIPS2020_fd69dbe2}: Apply the meta-learning algorithm to the initially given domains.
\end{itemize}

\begin{figure}[tbp]
\centering
  \includegraphics[clip, width=14cm]{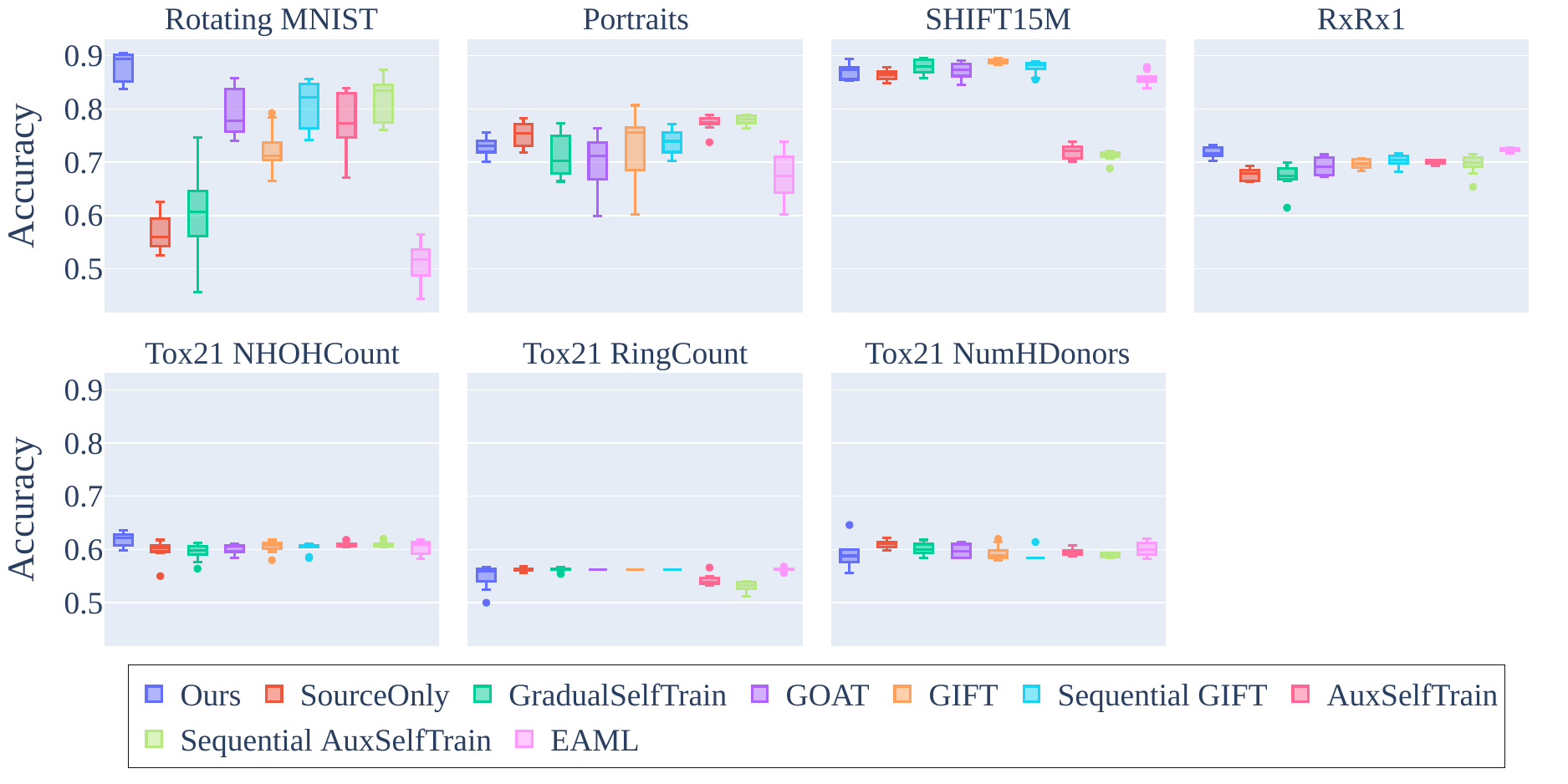}
  \caption{
  Comparison of accuracy on five real-world datasets. 
  }
  \label{fig:compare_baseline}
\end{figure}

Each evaluation was repeated 10 times using different initial weights of neural networks. 
We show the experimental results in Figure~\ref{fig:compare_baseline}. 
Our proposed method has comparable or superior accuracy to the baseline methods on all datasets. 
We consider a GDA problem with large discrepancies between adjacent domains. 
We estimate the distance between the source and target domains of each dataset by predicting the target dataset with SourceOnly.
The prediction performance of SourceOnly on {\tt Rotating MNIST} is low. 
It suggests there is a large gap between the source and target domains.
On the other hand, the gap is not as large for the other datasets. 
The proposed method is effective on datasets with a large gap between the source and target domains.
As mentioned in Section~\ref{sec:sub-exp-generate}, the prediction performance of a method that uses pseudo-intermediate domains and gradual self-training will deteriorate when suitable pseudo-intermediate domains for self-training are not obtained.
AuxSelfTrain is the only method among the baseline methods that incorporate unsupervised learning during self-training.
AuxSelfTrain seems to be suitable for the {\tt Portraits} dataset, but it does not appear to be suitable for the {\tt SHIFT15M} dataset.
EAML learns meta-representations from the sequence of unlabeled datasets.
In our problem setting, the number of given intermediate domains is limited, which may be insufficient for learning meta-representations.
The proposed method has demonstrated stable and comparable performance on all datasets. 
Moreover, as shown in Section~\ref{sec:sub-exp-generate}, our proposed method can generate artificial intermediate samples.

\section{Discussion and conclusion}\label{sec:disc&conc} 
Gradual domain adaptation is one of the promising approaches to addressing the problem of a large domain gap by leveraging the intermediate domain. 
\citet{kumar2020understanding} assumed in their previous work that the intermediate domain gradually shifts from the source domain to the target domain and that the distance between adjacent domains is small. 
In this study, we consider the problem of a large distance between adjacent domains. 
The proposed method mitigates the problem by utilizing normalizing flows, with a theoretical guarantee on the prediction error in the target domain. 
We evaluate the effectiveness of our proposed method on five real-world datasets. 
The proposed method mitigates the limit of the applicability of GDA. 

In this work, we assume that there is no noisy intermediate domain. 
A noisy intermediate domain deteriorates the predictive performance of the proposed method. 
It remains to be our future work to develop a method to select several appropriate intermediate domains from the given noisy intermediate domains. 

\section*{Acknowledgments}
Part of this work is supported by JSPS KAKENHI Grant No. JP22H03653, JST CREST Grant Nos. JPMJCR1761 and JPMJCR2015, JST-Mirai Program Grant No. JPMJMI19G1, and NEDO Grant No. JPNP18002.

\bibliography{ref}

\newpage
\appendix
\renewcommand{\thesection}{Appendix \Alph{section}}

\section{Experimental details}\label{sec:app-exp-detail}

\subsection{Networks}
We propose a gradual domain adaptation method that utilizes normalizing flows. 
The baseline methods include self-training-based methods and a method that uses meta-learning. 
The composition of the neural network for each method is shown as follows. 

\paragraph{Our proposed method}
One CNF block consists of two fully connected layers with 64 nodes in each layer. 
Our flow-based model $g$ consists of one CNF block. 

\paragraph{Self-training-based method}
Recall that these methods update the hypothesis $h: \mathcal{X} \to \mathcal{Y}$ trained on the source dataset by applying sequential self-training. 
We follow the hypothesis used by~\citet{he2023gradual} since they consider the same problem settings as ours. 
The hypothesis $h$ consists of an encoder and a classifier. 
The encoder has two convolutional layers, and the classifier has two convolutional layers and two fully connected layers. 
Since we apply UMAP to all datasets as preprocessing, we modify the convolutional layer of the model to fully connected layers. 
We use this hypothesis in all baseline methods except for the proposed method and EAML.

\paragraph{EAML}
EAML requires a feature extractor and a meta-adapter. 
In~\citet{NEURIPS2020_fd69dbe2}, the feature extractor consists of two convolutional layers, and the meta-adapter comprises two fully connected layers. 
We modify the convolutional layers of the feature extractor to fully connected layers since we apply UMAP to all datasets as preprocessing. 
Other necessary parameters during training are set as specified in~\citet{NEURIPS2020_fd69dbe2}.

\subsection{Datasets}
We use benchmark datasets with modifications for gradual domain adaptation. 
Since we are considering the situation that the distances of the adjacent domains are large, we prepare only one or two intermediate domains. 
We describe the details of the datasets.

\paragraph{{\tt Two Moon}}
A toy dataset. 
We use the {\tt two-moon} dataset as the source domain. 
The intermediate and target domains are prepared by rotating the source dataset by $\pi/4$ and $\pi/2$, respectively. 

\paragraph{{\tt Block}}
A toy dataset. 
The number of dimensions of the data is two, and the number of classes is five. 
We prepare the intermediate and target domains by adding horizontal movement to each class. 
Note that only the {\tt Block} dataset has two intermediate domains. 

\paragraph{{\tt Rotating MNIST}~\citep{kumar2020understanding}}
We add rotations to the {\tt MNIST} data. 
The rotation angle is $0$ for the source domain, $\pi/6$ for the intermediate domain, and $\pi/3$ for the target domain. 
We normalize the image intensity to the range between 0 and 1 by dividing by 255. 

\paragraph{{\tt Portraits}~\citep{ginosar2015century}}
The {\tt Portraits} dataset includes photographs of U.S. high school students from 1905 to 2013, and the task is gender classification. 
We sort the dataset in ascending order by year and split the dataset. 
The source dataset includes data from the 1900s to the 1930s. 
The intermediate dataset includes data from the 1940s and the 1950s. 
The target dataset includes data from the 1960s. 
We resize the original image to 32 x 32 and normalize the image intensity to the range between 0 and 1 by dividing by 255. 

\paragraph{{\tt Tox21}~\citep{thomas2018us}}
The {\tt Tox21} dataset contains the results of measuring the toxicity of compounds. 
The dataset contains 12 types of toxicity evaluation with a number of missing values. 
We merge these evaluations into a single evaluation and consider a compound as toxic when it is determined to be harmful in any of the 12 evaluations. 
Since {\tt Tox21} has no domain indicator such as year, we introduce an indicator for splitting the entire dataset into domains. 
It is a reasonable method from the chemical view point to divide the entire dataset into domains by the number of arbitrary substituents in the compound. 
We select the following three chemically representative substituents and use the number of substituents as a domain indicator.
\begin{itemize}
    \setlength{\parskip}{0cm}
    \setlength{\itemsep}{0.1cm}
   \item NHOHCount: Number of NHOH groups in the compound. 
   \item RingCount: Number of ring structures in the compound. 
   \item NumHDonors: Number of positively polarized hydrogen bonds in the compound. 
\end{itemize}
The zeroth, first, and second substituents are assigned to the source domain, the intermediate domain, and the target domain, respectively. 
We use 108-dimensional molecular descriptors as features used in the previous work~\citep{drwal2015molecular}.

\paragraph{{\tt SHIFT15M}~\citep{kimura2021shift15m}}
{\tt SHIFT15M} consists of 15 million fashion images collected from real fashion e-commerce sites. 
We estimate seven categories of clothes from image features. 
{\tt SHIFT15M} does not provide images but provides VGG16~\citep{Simonyan15} features consisting of 4,096 dimensions. 
The dataset contains fashion images from 2010 to 2020, and the passage of years causes a domain shift. 
The number of samples from 2010 is significantly smaller than that from other years, and we merge samples from 2010 with those from 2011. 
We consider the datasets from 2011, 2015, and 2020 as the source domain, the intermediate domain, and the target domain, respectively. 
Owing to the significant number of samples, we randomly select 5,000 samples from each domain. 

\paragraph{{\tt RxRx1}~\citep{taylor2019rxrx1}}
{\tt RxRx1} consists of three channels of cell images obtained by a fluorescence microscope. 
We resize the original image to 32 x 32 and normalize the image intensity to the range between 0 and 1 by dividing by 255. 
Domain shifts occur in the execution of each batch due to slight changes in temperature, humidity, and reagent concentration. 
We estimate the cell type used in the experiment from image features. 
We consider batch numbers one, two, and three as the source domain, the intermediate domain, and the target domain, respectively.

\subsection{Image generation}
We proposed a method that uses continuous normalizing flows (CNFs) to learn gradual shifts between domains. 
Our main purpose is gradual domain adaptation, but the trained CNFs can generate pseudo intermediate domains such as morphing as a byproduct of the use of normalizing flow. 
It is difficult to handle high-dimensional data directly due to the high-computational cost of off-the-shelf CNF implementation. 
Therefore, we showed the demonstration of image generation by combining the proposed method and variational autoencoder (VAE) in Figure~\ref{fig:vae} of the main body of the paper. 
Here, we denote the details of the experiment.

We assign 60,000 samples drawn from the {\tt MNIST} dataset without any rotation to the source domain, and the target domain is prepared by rotating images of the source dataset by angle $\pi/3$. 
Following~\citet{kumar2020understanding}, the intermediate domains are prepared by adding gradual rotations, angles from $\pi/84$ to $\pi/3.11$. 
The number of intermediate domains is 27 in total. 
To obtain the latent variables that shift the source domain to the target domain gradually, we use all the intermediate domains for the training of VAE. 
After the training of VAE, we extract the latent variables whose rotation angles correspond to $0$, $\pi/6$, and $\pi/3$. 
Following~\citet{kumar2020understanding}, we randomly select 2,000 samples from each domain. 
We train our flow-based model with the latent variables and generate pseudo intermediate domains by using trained CNF. 
Figure~\ref{fig:vae} of the main body of the paper shows images from decoded pseudo intermediate domains.

\newpage
\section{Proofs}\label{sec:app-proof} 
Here, we show the details of proofs. 
Note that Propositions~\ref{prop:kl-bound} and~\ref{prop:cnf} were originally derived by~\citet{nguyen2022kl} and~\citet{onken2021ot}, respectively. 
For the sake of completeness, we provide proofs using the notations consistent with those used in this paper. 

\subsection{Proof of proposition~\ref{prop:kl-bound}~\citep{nguyen2022kl}}
\begin{proof}
Recall that the expected losses on the source and $j$-th domains are defined as 
$\mathrm{L_1} = \mathbb{E}_{\bm{x},y \sim p_1(\bm{x},y)}[-\log p(y|\bm{x})]$ and 
$\mathrm{L_j} = \mathbb{E}_{\bm{x},y \sim p_j(\bm{x},y)}[-\log p(y|\bm{x})]$, respectively. 
In the standard domain adaptation, there is no intermediate domain and $K = 2$. We have
\begin{align}
    \mathrm{L_2} 
    &= \mathbb{E}_{p_2(\bm{x},y)}[-\log p(y|\bm{x})] \\
    &= \int - [\log p(y|\bm{x})] p_2(\bm{x},y) d\bm{x}dy \\
    &= \int - [\log p(y|\bm{x})] p_1(\bm{x},y) d\bm{x}dy + \int -\log p(y|\bm{x})[p_2(\bm{x},y)-p_1(\bm{x},y)] d\bm{x}dy \\
    &= \mathrm{L_1} + \int -\log p(y|\bm{x})[p_2(\bm{x},y)-p_1(\bm{x},y)] d\bm{x}dy. \label{eq:disc}
\end{align}
We define sets $\mathcal{A}$ and $\mathcal{B}$ as 
\begin{align}
    \mathcal{A} = \{(\bm{x},y)|p_2(\bm{x},y)-p_1(\bm{x},y) \ge 0\}, \hspace{3mm}
    \mathcal{B} = \{(\bm{x},y)|p_2(\bm{x},y)-p_1(\bm{x},y) < 0\}.
\end{align}
If Assumption~\ref{ass:logp} holds, we have
\begin{align}
    & \int - \log p(y|\bm{x})[p_2(\bm{x},y)-p_1(\bm{x},y)] d\bm{x}dy \\
    &= \int_\mathcal{A} - \log p(y|\bm{x})[p_2(\bm{x},y)-p_1(\bm{x},y)] d\bm{x}dy \!+\! \int_\mathcal{B} -\log p(y|\bm{x})[p_2(\bm{x},y)-p_1(\bm{x},y)] d\bm{x}dy \\
    &\leq \int_\mathcal{A} -\log p(y|\bm{x})[p_2(\bm{x},y)-p_1(\bm{x},y)] d\bm{x}dy \\
    &= \int_\mathcal{A} -\log p(y|\bm{x})|p_2(\bm{x},y)-p_1(\bm{x},y)| d\bm{x}dy \\
    &\leq M \int_\mathcal{A} |p_2(\bm{x},y)-p_1(\bm{x},y)| d\bm{x}dy \hspace{5mm} (\; \because -\log p(y|\bm{x}) \leq M),
\end{align}
where $|\cdot|$ is the absolute value.
Note that $\int_\mathcal{A} |p_2(\bm{x},y)-p_1(\bm{x},y)| d\bm{x}dy$ is called the total variation of two distributions.
From the identity $
\int p_2(\bm{x},y)-p_1(\bm{x},y) d\bm{x}dy = 0$, we have
\begin{align}
&\int_\mathcal{A} p_2(\bm{x},y)-p_1(\bm{x},y) d\bm{x}dy + \int_\mathcal{B} p_2(\bm{x},y)-p_1(\bm{x},y) d\bm{x}dy = 0 \\
    &\Leftrightarrow \int_\mathcal{A} p_2(\bm{x},y)-p_1(\bm{x},y) d\bm{x}dy = \int_\mathcal{B} p_1(\bm{x},y) - p_2(\bm{x},y) d\bm{x}dy \\
    &\Leftrightarrow \int_\mathcal{A} |p_2(\bm{x},y)-p_1(\bm{x},y)| d\bm{x}dy = \int_\mathcal{B} |p_2(\bm{x},y)-p_1(\bm{x},y)| d\bm{x}dy \\
    &\Leftrightarrow \int_\mathcal{A} |p_2(\bm{x},y)-p_1(\bm{x},y)| d\bm{x}dy = \frac{1}{2} \int |p_2(\bm{x},y)-p_1(\bm{x},y)| d\bm{x}dy.
\end{align}
Therefore,
\begin{align}
    \mathrm{L_2} 
    &= \mathrm{L_1} + \int -\log p(y|\bm{x})[p_2(\bm{x},y)-p_1(\bm{x},y)] d\bm{x}dy \\
    &\leq \mathrm{L_1} + M \int_\mathcal{A} |p_2(\bm{x},y)-p_1(\bm{x},y)| d\bm{x}dy \\
    &= \mathrm{L_1} + \frac{M}{2} \int |p_2(\bm{x},y)-p_1(\bm{x},y)| d\bm{x}dy.
\end{align}
Using the Pinsker's inequality, we have
\begin{equation}
    \left(\int |p_2(\bm{x},y)-p_1(\bm{x},y)| d\bm{x}dy\right)^2 \leq 2 \int p_2(\bm{x},y)\log\frac{p_2(x,y)}{p_1(x,y)} d\bm{x}dy.
\end{equation}
Therefore,
\begin{align}
    \mathrm{L_2} 
    &\leq \mathrm{L_1} + \frac{M}{2} \sqrt{2 \int p_2(\bm{x},y)\log\frac{p_2(x,y)}{p_1(x,y)} d\bm{x}dy} \\
    &= \mathrm{L_1} + \frac{M}{\sqrt{2}} \sqrt{\mathrm{KL}[p_2(\bm{x},y)|p_1(\bm{x},y)]}.
\end{align}
We decompose the KL divergence between $p_2(\bm{x},y)$ and $p_1(\bm{x},y)$ into the marginal and conditional misalignment terms as follows:
\begin{align}
    &\mathrm{KL}[p_2(\bm{x},y)|p_1(\bm{x},y)] \\
    &= \mathbb{E}_{p_2(\bm{x},y)}[\log p_2(\bm{x},y)-\log p_1(\bm{x},y)] \\
    &= \mathbb{E}_{p_2(\bm{x},y)}[\log p_2(\bm{x}) + \log p_2(y|\bm{x}) - \log p_1(\bm{x}) - \log p_1(y|\bm{x})] \\
    &= \mathbb{E}_{p_2(\bm{x},y)}[\log p_2(\bm{x}) - \log p_1(\bm{x})] + \mathbb{E}_{p_2(\bm{x},y)}[\log p_2(y|\bm{x}) - \log p_1(y|\bm{x})] \\
    &= \mathrm{KL}[p_2(\bm{x})|p_1(\bm{x})] + \mathbb{E}_{p_2(\bm{x})}[\mathbb{E}_{p_2(y|\bm{x})}[\log p_2(y|\bm{x}) - \log p_1(y|\bm{x})]] \\
    &= \mathrm{KL}[p_2(\bm{x})|p_1(\bm{x})] + \mathbb{E}_{p_2(\bm{x})}[\mathrm{KL}[p_2(y|\bm{x})|p_1(y|\bm{x})]].
\end{align}
Therefore, we have
\begin{align}
    \mathrm{L_2} 
    &\leq \mathrm{L_1} + \frac{M}{\sqrt{2}} \sqrt{\mathrm{KL}[p_2(\bm{x})|p_1(\bm{x})] + \mathbb{E}_{p_2(\bm{x})}[\mathrm{KL}[p_2(y|\bm{x})|p_1(y|\bm{x})]]},
\end{align}
which completes the proof.
\end{proof}

\subsection{Proof of proposition~\ref{prop:cnf}~\citep{onken2021ot}}
\begin{proof}
Let $p_{t+1}$ be the initial density of the samples $\bm{x} \in \mathbb{R}^d$ and $g: \mathbb{R}^{d} \times \mathbb{R}_{+} \to \mathbb{R}^{d}$ be the trajectories that transform samples from $p_{t+1}$ to $p_{t}$.
The change in density when $p_{t+1}$ is transformed from time $t+1$ to $t$ is given by the change of variables formula
\begin{equation}
    p_{t+1}(\bm{x}) = p_{t+1}^{\ast}(g(\bm{x},t))|\mathrm{det} \nabla g(\bm{x},t)|,
    \label{eq:cvf}
\end{equation}
where $p_{t+1}^{\ast}$ and $\nabla g(\bm{x},t)$ are the transformed density and the Jacobian of $g$, respectively. 
Normalizing flows aim to learn a function $g$ that transforms $p_{t+1}$ to $p_t$. 
Measuring the discrepancy between the transformed and objective distributions indicates whether the trained function $g$ is appropriate. 
The discrepancy between two distributions is measured using the KL divergence
\begin{equation}
    \mathrm{KL}[p_{t+1}^{\ast}(\bm{x}) | p_{t}(\bm{x})] 
    = \int_{\mathbb{R}^d} \log \left(\frac{p_{t+1}^{\ast}(\bm{x})}{p_{t}(\bm{x})}\right) p_{t+1}^{\ast}(\bm{x}) d\bm{x}.
    \label{eq:cnf-kl}
\end{equation}
We transform a sample $\bm{x}$ using $g$. 
By using the change of variable formula, we rewrite Eq.~\eqref{eq:cnf-kl} as follows
\begin{align}
    \mathrm{KL}[p_{t+1}^{\ast}(\bm{x}) | p_{t}(\bm{x})] 
    &= \int_{\mathbb{R}^d} \log \left(\frac{p_{t+1}^{\ast}(\bm{x})}{p_{t}(\bm{x})}\right) p_{t+1}^{\ast}(\bm{x}) d\bm{x} \\
    &= \int_{\mathbb{R}^d} \log \left(\frac{p_{t+1}^{\ast}(g(\bm{x},t))\bcancel{|\mathrm{det} \nabla g(\bm{x},t)|}}{p_{t}(g(\bm{x},t))\bcancel{|\mathrm{det} \nabla g(\bm{x},t)|}}\right) p_{t+1}^{\ast}(g(\bm{x},t))|\mathrm{det} \nabla g(\bm{x},t)| d\bm{x} \\
    &= \int_{\mathbb{R}^d} \log \left(\frac{p_{t+1}^{\ast}(g(\bm{x},t))}{p_{t}(g(\bm{x},t))}\right) p_{t+1}^{\ast}(g(\bm{x},t))|\mathrm{det} \nabla g(\bm{x},t)| d\bm{x} \\
    &= \int_{\mathbb{R}^d} \log \left(\frac{p_{t+1}^{\ast}(g(\bm{x},t))}{p_{t}(g(\bm{x},t))}\right) p_{t+1}(\bm{x}) d\bm{x}
    \hspace{5mm} (\because \; Eq.~\eqref{eq:cvf}).
\end{align}
By using Eq.~\eqref{eq:cvf}, we have
\begin{align}
    \mathrm{KL}[p_{t+1}^{\ast}(\bm{x}) | p_{t}(\bm{x})] 
    &= \int_{\mathbb{R}^d} \log \left(\frac{p_{t+1}^{\ast}(g(\bm{x},t))}{p_{t}(g(\bm{x},t))}\right) p_{t+1}(\bm{x}) d\bm{x} \\
    &= \int_{\mathbb{R}^d} \log \left( \frac{p_{t+1}(\bm{x})}{p_{t}(g(\bm{x},t))|\mathrm{det} \nabla g(\bm{x},t)|} \right) p_{t+1}(\bm{x}) d\bm{x} \\
    &= \mathbb{E}_{p_{t+1}(\bm{x})}\left[ \log p_{t+1}(\bm{x}) - \{\log p_{t}(g(\bm{x},t)) + \log |\mathrm{det} \nabla g(\bm{x},t)| \}\right].
    \label{eq:proof-int-kl}
\end{align}
\citet{chen2018neural} proposed neural ordinary differential equations, in which a neural network $v$ parametrized by $\omega$ represents the time derivative of the function $g$, as follows
\begin{equation}
    \frac{\partial g}{\partial v} = v(g(\cdot, t), t; \omega).
\end{equation}
Moreover, they showed that the instantaneous change of the density can be computed as follows
\begin{equation}
    \frac{\partial \log p(g)}{\partial t} = -\mathrm{Tr}\left(\frac{\partial v}{\partial g}\right).
\end{equation}
Namely, the term $\log |\mathrm{det} \nabla g(\bm{x},t)|$ in Eq.~\eqref{eq:proof-int-kl} is equivalent to $-\int_{t}^{t+1} \mathrm{Tr}(\partial v / \partial g)dt$.
Therefore, we rewrite the Eq.~\eqref{eq:proof-int-kl} as follows:
\begin{align}
    \mathrm{KL}[p_{t+1}^{\ast}(\bm{x}) | p_{t}(\bm{x})]
    &= \mathbb{E}_{p_{t+1}(\bm{x})}\left[ \log p_{t+1}(\bm{x}) - \left\{\log p_{t}(g(\bm{x},t)) -\int_{t}^{t+1} \mathrm{Tr}\left(\frac{\partial v}{\partial g}\right)dt\right\}\right]
    \label{eq:expected-tr}
\end{align}
Recall that the log-likelihood of continuous normalizing flow is given by
\begin{equation}
    \log p_{t+1}(g(\bm{x},t+1)) 
    \!=\! \log p_{t}(g(\bm{x},t)) 
    - \int_{t}^{t+1} \mathrm{Tr} \left(\frac{\partial v}{\partial g}\right) dt.
\end{equation}
Therefore, we have
\begin{align}
    \mathrm{KL}[p_{t+1}^{\ast}(\bm{x}) | p_{t}(\bm{x})]
    &= \mathbb{E}_{p_{t+1}(\bm{x})}\left[ \log p_{t+1}(\bm{x}) - \log p_{t+1}(g(\bm{x}, t+1))\right].
    \label{eq:min-kl}
\end{align}
The first term of Eq.~\eqref{eq:min-kl} does not depend on CNF $g$, and we can ignore it during the training of the CNF. 
Therefore, the minimization of Eq.~\eqref{eq:cnf-expected} is equivalent to the minimization of the KL divergence between $p_{t}$ and $p_{t+1}$ transformed by CNF $g$. 
\end{proof}

\subsection{Proof of corollary~\ref{corollary}}
\begin{proof}
In gradual domain adaptation, since the ordered sequence of the intermediate domains is given, we extend Eq.~\eqref{eq:kl-bound-logp} until the target loss $\mathrm{L_K}$ as follows: 
\begin{align}
    \mathrm{L_2} &\leq \mathrm{L_1} + \frac{M}{\sqrt{2}} \sqrt{\mathrm{KL}[p_2^{\ast}(\bm{x})|p_1(\bm{x})] + \mathbb{E}_{p_2(\bm{x})}[\mathrm{KL}[p_2(y|\bm{x})|p_1(y|\bm{x})]]} \\
    \mathrm{L_3} &\leq \mathrm{L_2} + \frac{M}{\sqrt{2}} \sqrt{\mathrm{KL}[p_3^{\ast}(\bm{x})|p_2(\bm{x})] + \mathbb{E}_{p_3(\bm{x})}[\mathrm{KL}[p_3(y|\bm{x})|p_2(y|\bm{x})]]} \\
    \vdots \\
    \mathrm{L_K} &\leq \mathrm{L_{K-1}} + \frac{M}{\sqrt{2}} \sqrt{\mathrm{KL}[p_K^{\ast}(\bm{x})|p_{K-1}(\bm{x})] + \mathbb{E}_{p_K(\bm{x})}[\mathrm{KL}[p_K(y|\bm{x})|p_{K-1}(y|\bm{x})]]}.
\end{align}
Summing up both sides of the above inequalities, we have
\begin{equation}
    \mathrm{L_K} 
    \leq \mathrm{L_1} + \frac{M}{\sqrt{2}} \sum_{t=2}^{K} \sqrt{\mathrm{KL}[p_t^{\ast}(\bm{x})|p_{t-1}(\bm{x})] + \mathbb{E}_{p_t(\bm{x})}[\mathrm{KL}[p_t(y|\bm{x})|p_{t-1}(y|\bm{x})]]}.
\end{equation}
If Assumption~\ref{ass:covariate} holds, we have
\begin{equation}
    \mathrm{L_K} 
    \leq \mathrm{L_1} + \frac{M}{\sqrt{2}} \sum_{t=2}^{K} \sqrt{\mathrm{KL}[p_t^{\ast}(\bm{x})|p_{t-1}(\bm{x})]}.
\end{equation}
\end{proof}

\end{document}